\documentclass[numbers,square,sort&compress]{assets/matterlab}

\usepackage{microtype}
\usepackage{graphicx}
\usepackage{booktabs}
\usepackage{float}
\usepackage{xurl}
\usepackage{hyperref}
\usepackage{titlesec}

\usepackage{amsmath}
\usepackage{amssymb}
\usepackage{mathtools}
\usepackage{amsthm}
\usepackage{bm}

\usepackage[noabbrev,nameinlink]{cleveref}

\usepackage{amsmath,amsfonts,bm}

\def\eqref#1{equation~\ref{#1}}

\def\1{\bm{1}}

\DeclareMathAlphabet{\mathsfit}{\encodingdefault}{\sfdefault}{m}{sl}
\SetMathAlphabet{\mathsfit}{bold}{\encodingdefault}{\sfdefault}{bx}{n}

\usepackage{orcidlink}
\usepackage{comment}
\usepackage[version=4]{mhchem}
\usepackage{longtable}
\usepackage{array}

\usepackage{multirow}           %
\usepackage{amsfonts}           %
\usepackage{nicefrac}
\usepackage{duckuments}         %
\usepackage{thmtools,thm-restate}
\usepackage{enumitem}
\usepackage{xcolor,colortbl}
\usepackage{tikz}
\usepackage{tikz-cd}
\usepackage{caption}
\usepackage{subcaption}
\usepackage{listings}
\usepackage{gensymb}
\usepackage{textcomp} %
\usepackage[inkscapelatex=false]{svg}

\providecommand{\doi}[1]{\href{https://doi.org/#1}{doi:#1}}

\usepackage{cinzel}
\newcommand{\dmta}[1]{\texorpdfstring{{\cinzel #1}}{#1}}
\setcitestyle{numbers,square,comma}

\title{Closing the Analyze-to-Design Gap: An Authorization-Bounded Agent for Self-Driving Laboratories}

\author[1,2]{Yimeng Liu}
\author[2,\ddagger]{Eric Isbrandt}
\author[3,\ddagger]{Felix Katzenburg}
\author[2,4]{Kevin P. Greenman}
\author[2]{Han Hao}
\author[2]{Yang Cao}
\author[1,2,3,*]{Varinia Bernales}
\author[1,2,3,5,6,7,8,9,10,*]{Al\'an Aspuru-Guzik}

\affiliation[1]{Department of Computer Science, University of Toronto, 40 St George St., Toronto, ON M5S 2E4, Canada}
\affiliation[2]{Acceleration Consortium, 700 University Ave., Toronto, ON M7A 2S4, Canada}
\affiliation[3]{Department of Chemistry, University of Toronto, 80 St. George St., Toronto, ON M5S 3H6, Canada}
\affiliation[4]{Catholic Institute of Technology, 1 Broadway, 14th Floor, Cambridge, MA, United States of America}
\affiliation[5]{Department of Materials Science \& Engineering, University of Toronto, 184 College St., Toronto, ON M5S 3E4, Canada}
\affiliation[6]{Department of Chemical Engineering \& Applied Chemistry, University of Toronto, 200 College St., Toronto, ON M5S 3E5, Canada}
\affiliation[7]{Institute of Medical Science, 1 King's College Circle, Medical Sciences Building, Room 2374, Toronto, ON M5S 1A8, Canada}
\affiliation[8]{Vector Institute for Artificial Intelligence, W1140-108 College St., Schwartz Reisman Innovation Campus, Toronto, ON M5G 0C6, Canada}
\affiliation[9]{NVIDIA, 431 King St W \#6th, Toronto, ON M5V 1K4, Canada}
\affiliation[10]{Canadian Institute for Advanced Research (CIFAR), 661 University Ave., Toronto, ON M5G 1M1, Canada}

\contribution[\ddagger]{These authors contributed equally as second authors.}

\abstract{
Artificial intelligence (AI) agents are increasingly integrated into self-driving laboratories (SDLs) to plan experiments and interpret results. However, analytical reasoning alone does not make a laboratory autonomous: autonomy in an SDL is achievable only when the agent can be trusted to act on physical hardware unsupervised and knows when to defer to a human. Here, we present an architecture that integrates an agent at the \dmta{A}nalyze--\dmta{D}esign interface of the Design--Make--Test--Analyze (DMTA) cycle, enabling agent-driven measurement selection and data interpretation while ensuring that state-changing operations are authorized, validated, and auditable. We demonstrate this architecture on a liquid chromatography--mass spectrometry (LC--MS) instrument in an oxidant-screening campaign for synthesizing an organic dye. Within human-defined campaign limits, the agent submitted measurement jobs, interpreted mass spectrometry (MS) and diode-array detection (DAD) data, flagged failed experiments, and produced an oxidant ranking. This ranking matched that obtained from scripted peak integration, and organic chemists confirmed the agent’s spectral interpretations and failure assignments. In a simulated environment that replays the archived LC–MS data, where the agent chose which wells to inject under varying injection limits, it measured the highest-yielding condition at about its seventh injection, against 14.5 on average for a random order, and identified it in every campaign with a limit of eight or more injections. An agent can therefore close the analysis-to-design gap while its actions stay bounded and verifiable, which is a step toward greater autonomy for SDLs.

}

\date{\today}
\correspondence{\email{alan@aspuru.com} and \email{varinia@bernales.org}}

\begin{document}

\maketitle

\section{Introduction}\label{sec:intro}

Self-driving laboratories (SDLs) are evolving from execution-centered automation toward agent-mediated systems \citep{Has19,Tom24,Asp26}. Yet translating experimental results into the next experimental design remains difficult to automate \citep{Sch25}. These decisions often require adapting to unexpected results and weighing scientific context, capabilities that rigid rule-based systems struggle to accommodate. At the same time, many such decisions involve recurring judgments that require scientific expertise but must be made repeatedly between experimental iterations. This creates an opportunity for agentic systems to take on these recurring decisions, bringing the flexibility needed to adapt experimental strategies as results emerge. However, flexibility alone is not enough: agents must remain bounded by structured protocols, experimental state, safety controls \citep{Leo25,Kang26,Asp26}, provenance, and expert oversight \citep{Lee26,Xie26,Che26,Zha26b}. Recent systems address individual aspects of this problem through schema-gated orchestration \citep{Str26}, deterministic tool mediation \citep{Ada26}, and verification or simulation of generated protocols before execution \citep{Yos23,Pag26}. What remains less clear is how these mechanisms should be integrated into existing laboratory setups and SDL architecture: where the agent should sit, what authority it should receive, and how its decisions should be validated and traced \citep{Xie26,Che26,Leo25,Asp26}.

An SDL typically runs through repeated Design--Make--Test--Analyze (DMTA) cycles: researchers design experiments, make samples, test them, and analyze the resulting measurements \citep{Has19,Tom24}. Prior SDL work has advanced the automation of making and testing \citep{Has19,Tom24}. To move beyond task automation toward greater autonomy in scientific discovery, we focus on connecting analysis with experiment design rather than treating design as an isolated capability. We argue that useful hypotheses should emerge from interpreting accumulated experimental results and guide the next experiment. This analysis-to-design (\dmta{A}-to-\dmta{D}) handoff is more than data processing: it requires interpreting results in light of prior experiments, campaign objectives, and practical constraints, then selecting what to measure or make next. The key architecture question is how an agent can support this handoff while remaining within the laboratory’s safety and authority boundaries.

\begin{figure}[H]
\centering
\includegraphics[width=0.5\textwidth]{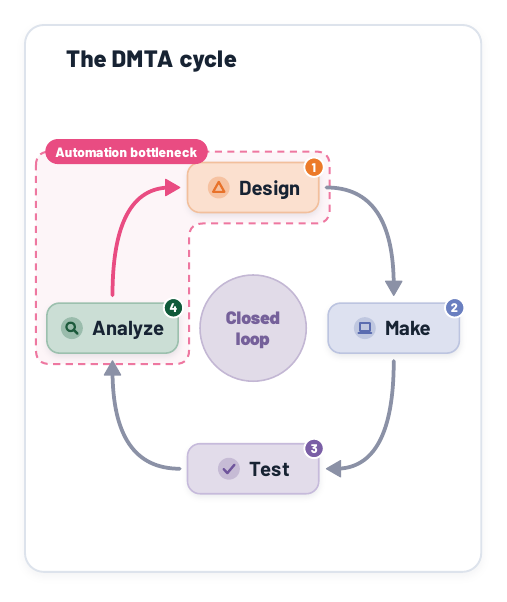}
\caption{The DMTA cycle. The step from \dmta{A}nalyze to \dmta{D}esign, which today needs a human in every iteration, is the one that this work addresses.}
\label{fig:dmta}
\end{figure}

We present an SDL-native architecture focused on the \dmta{A}nalysis and \dmta{D}esign stages of the DMTA cycle (Figure~\ref{fig:dmta}). The agent accesses scientific and experimental context to propose and interpret next actions, but laboratory state can only be changed through validated, permission-scoped, expert-controlled interfaces. A shared human--AI workspace keeps the agent's reasoning and results visible, while deterministic analysis, workflow validation, campaign limits, and auditable execution traces constrain its actions.

Before a campaign, the expert provides its requirements, from which the agent drafts operational limits covering the objective, metric, instrument method, injection volume and budget, and allowed wells. The expert reviews and approves these limits and separately enables agent execution. Once authorized, the agent can select, submit, and interpret measurements within the approved limits without per-iteration sign-off. The expert retains the ability to revoke execution or stop the agent at any time. This shifts human intervention from routine experimental iterations to exceptional decisions requiring expert judgment.

\begin{figure}[!htb]
\centering
\includegraphics[width=\textwidth]{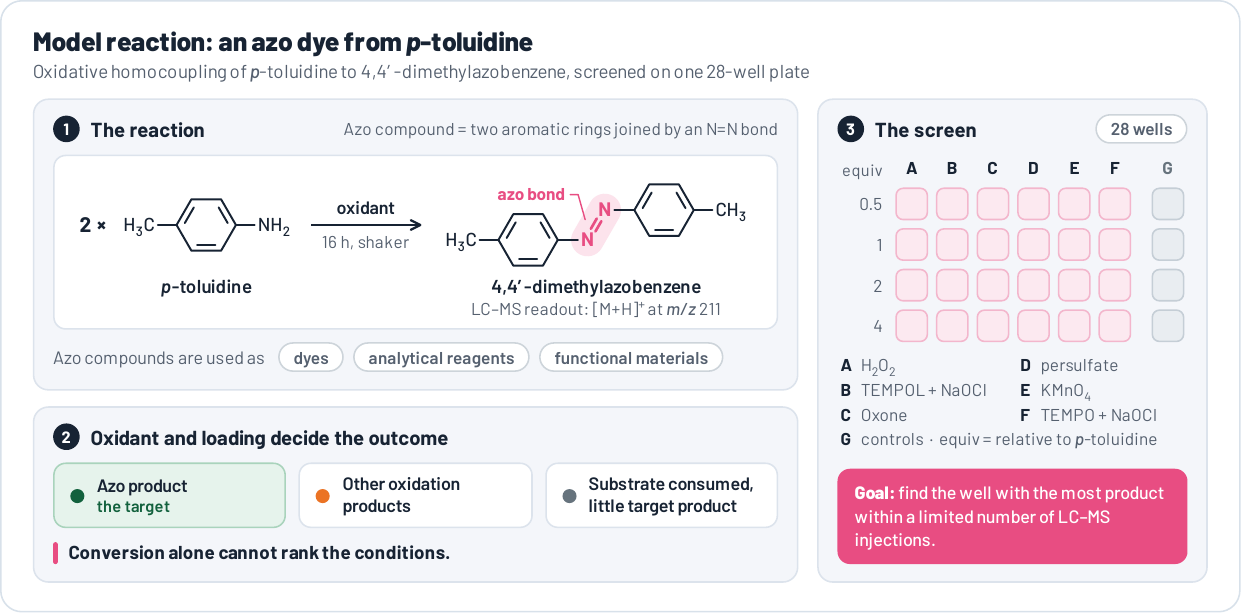}
\caption{The benchmark reaction and the screen. (1) Oxidative homocoupling of \textit{p}-toluidine gives the azo compound 4,4$^\prime$-dimethylazobenzene, detected by LC--MS. (2) Depending on the oxidant and its loading, the substrate can instead give other oxidation products or be consumed with little target product, so conversion alone cannot rank the conditions. (3) Six oxidant systems at four loadings, plus four controls, on one 28-well plate.}
\label{fig:azo}
\end{figure}

To evaluate the architecture, we selected the oxidative homocoupling of p-toluidine to 4,4'-dimethylazobenzene (Figure~\ref{fig:azo}) as a benchmark that is chemically relevant and analytically demanding. The reaction itself is known. Azo compounds, in which two aromatic rings are joined by an \ce{N=N} bond, have applications in dyes, analytical reagents, and functional materials, and oxidative coupling offers an alternative to conventional diazotization-and-coupling routes \citep{eltaboni2022azo,Geo15}. Prior work has explored the oxidation of p-toluidine and related aniline derivatives \citep{Wu08,Geo15}, but, to our knowledge, this specific screen has not been conducted in an SDL.

The reaction provides a useful test because varying the oxidant and loading can produce the target azo compound, other oxidation products, or substrate consumption with little target product. Conversion alone is therefore insufficient to rank conditions; the agent must interpret mass spectrometry (MS) and diode-array detection (DAD) evidence across wells. During the physical campaign, the agent submitted liquid chromatography--mass spectrometry (LC--MS) measurements to a real instrument, interpreted results, flagged failed sample transfers, and ranked conditions. Because the operator extended the injection budget to cover every well, we separately replayed archived data under fixed budgets to evaluate well selection. The study demonstrates a bounded agent--laboratory workflow, and Section~\ref{sec:limitations} states what it does not show.

\section{Related Work}\label{sec:related}

\subsection{Chemical automation and executable procedures}

Self-driving laboratories in chemistry and materials build on a long history of laboratory automation and high-throughput experimentation \citep{Gey84,Dev14}. Efforts to advance laboratory automation have focused on executable protocols, portable instrument control, and adaptive experimentation. LeyLab combined remote reaction monitoring and instrument control with autonomous self-optimization \citep{Fit15}. The Chemputer introduced modular robotic synthesis through a chemical programming language \citep{Ste19}, followed by XDL, which made literature procedures machine-readable and executable \citep{Meh20}. Subsequent work extended these approaches through versioned validation records and cross-platform protocol execution \citep{Roh22,Rau24}, as well as reusable reaction blueprints and dependency-aware parallel workflows \citep{Sia24}. In parallel, ARChemist explored the reconfiguration of heterogeneous laboratory equipment around sample-aware workflows \citep{Fak22}. Despite this progress, practical deployment remains chemistry-specific: integrating instruments, procedures, and data requires substantial iterative engineering beyond the automation of written protocols \citep{Chr21}.

\subsection{Orchestration and closed-loop discovery}

SDL orchestration systems coordinate instruments, experimental state, data, and optimization. ChemOS provides a modular software layer for connecting automated equipment and managing experiments across different levels of researcher involvement, with ChemOS 2.0 extending this framework to distributed devices, databases, and workflow services \citep{Roc20,Sim24}. The distinction between fixed automation and goal-directed autonomy has also been examined in automated research platforms, particularly its implications for interfaces and scheduling \citep{Can23}. ESCALATE complements this infrastructure with machine-readable experiment specifications, metadata capture, and structured reporting for algorithm-guided chemistry \citep{Pen19}. Beyond infrastructure, closed-loop studies have integrated algorithmic experiment selection, robotic execution, and analytical feedback for reaction optimization and programmable synthesis \citep{Chr20,Ang22,Leo24}. Other work has extended discovery across distributed experiments through asynchronous optimization of organic laser emitters and knowledge graphs for sharing laboratory state and provenance \citep{Str24,Bai22,Bai24}. Together, these systems establish the orchestration and data infrastructure needed for autonomous experimentation, but leave open how agentic reasoning should be integrated into these workflows.

\subsection{Agentic chemistry and laboratory systems}

Language-model-based chemistry systems introduce flexible goal interpretation, information retrieval, tool use, and procedure generation. Early demonstrations explored tool-augmented chemical reasoning and autonomous research \citep{Boi23,Bra23}, followed by end-to-end synthesis development, multi-agent robotic experimentation, and literature-to-executable-procedure workflows \citep{Rua24,Son25,Pag26}. More recent systems extend agents into physical laboratory workflows, including agent-supervised optimization \citep{Mul26b}, autonomous purification \citep{Cai26}, and broader agentic chemistry platforms \citep{Xie26,Feh25d,Gao25,Ang26,Che26}. However, the extent of experimental autonomy varies: agent involvement does not necessarily entail autonomous sample preparation or end-to-end execution of the DMTA cycle.

\subsection{Execution boundaries and positioning}

Because agents can interpret ambiguous goals and revise plans, their proposed actions require a controlled path into laboratory execution. Existing approaches address different aspects of this challenge: schema-gated workflows separate conversational planning from authorized execution \citep{Str26}, deterministic tool mediation constrains how model outputs reach scientific services \citep{Ada26}, and robotics and chemputation systems validate generated procedures against task, domain, or hardware constraints \citep{Yos23,Pag26}. Other work emphasizes SDL safety, human oversight, interlocks, and auditable action records \citep{Leo25,Asp26,Zha26b,Kang26,munguia2026chemisteye}. However, how these safeguards should work together within a single laboratory architecture remains unclear. We address this gap at the \dmta{A}nalyze--\dmta{D}esign interface: the agent selects and interprets measurements within approved campaign limits, while laboratory services retain authority over instrument execution and experimental records.

\section{System Architecture and Implementation}\label{sec:architecture}
\begin{figure}[!htb]
\centering
\includegraphics[width=0.9\textwidth]{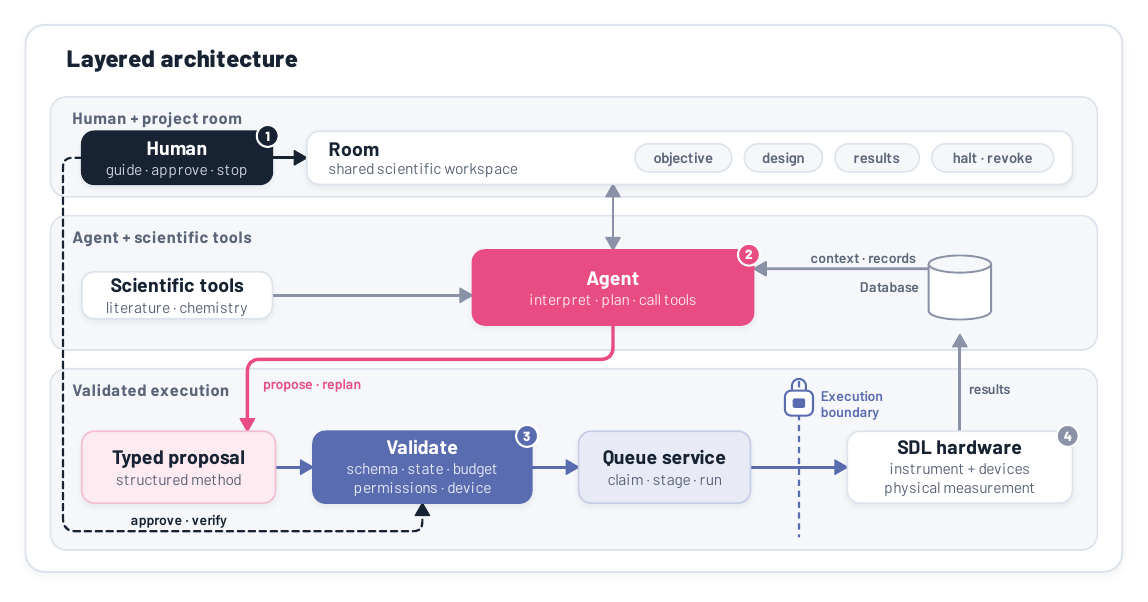}
\caption{Three-layer architecture. The researcher (1) guides and can stop the agent (2), which interprets results and plans experiments but cannot reach the instrument directly. Each proposed action must pass validation (3) before it runs on the SDL hardware (4).}
\label{fig:architecture}
\end{figure}
The implementation maps the analysis-to-design transition onto three operational layers (Figure~\ref{fig:architecture}): a human-supervised project room, an agent that interprets scientific and campaign context, and laboratory services that validate and execute approved requests. The key invariant is that the agent can propose a state-changing action but cannot bypass the services that authorize it (Appendix \ref{app:ui-agent}).

\subsection{Deployed system}\label{sec:deployed}
\begin{figure}[!htb]
    \centering
    \includegraphics[width=0.8\textwidth]{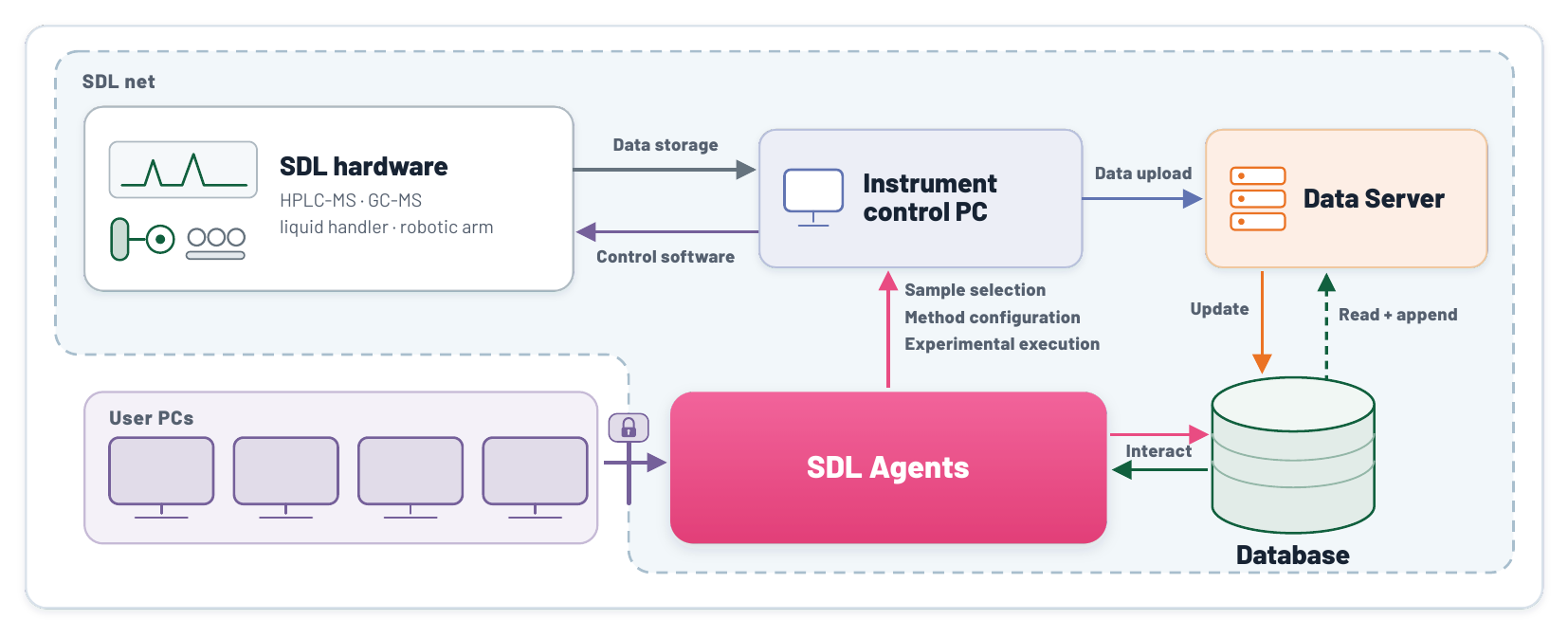}
    \caption{SDL Agents and their environments. Agent access is authenticated and permission-scoped, so each user reaches only the campaigns and records they are authorized to see.}
    \label{fig:network}
\end{figure}

Figure~\ref{fig:network} shows the deployed stack. An instrument-control PC runs the vendor chromatography data system together with MoSeS \citep{Had26}, an open-source Python framework from our group that provides sample-by-sample instrument control and analyte-resolved processing of chromatography--mass spectrometry data, so that measurements can be submitted and processed by software. Alongside it, a sidecar service provides status, cooperative claims, heartbeats, a first-in, first-out (FIFO) queue, and an approved-method allow-list. Runs are uploaded to a PostgreSQL experiment database that contains experiments, samples, measurements, files, and analyses. The agent service is the only route from the laboratory server to the instrument PC; it reads and writes experiment records and stores room state, tool calls, plans, and the campaign ledger in MongoDB. User PCs enter through an authenticating edge, and actions are attributed to the authenticated room owner. More details on the user interface are in Appendix \ref{app:workspace}. 

\subsection{Grounding, context, and the shared room}

The agent reads the objective, design, plate layout, measurements, analyses, constraints, and stopping criteria from project-scoped records, and grounds its reasoning in the literature through a search tool over OpenAlex, a free scholarly index, and through general web search. Its LC--MS tools, built on MoSeS, extract chromatograms, peaks, and peak-area ratios, and compute expected ion masses and isotope patterns from a molecular formula and adduct; cheminformatics libraries (RDKit and PubChemPy) are also available through a Python tool. The agent interprets these outputs and selects the next measurement.

Each project has a shared room for the researchers and the agent. The room holds a conversation window, an editable design, the campaign and instrument status, a visualization of the results, and a history log. A per-room \emph{agent-execution} toggle enables operation, while each campaign requires one-time approval of its objective, metric, method, injection volume, budget, and allowed wells. Each room also has an emergency stop that overrides both.

\subsection{Checking actions before agents reach the instrument}
\label{sec:checks}

By design, an agent proposal never becomes an instrument command, and the agent cannot bypass the checks below: each request is checked in sequence by the software, with failures returned as structured error messages.

\textbf{Agent process.} Tool arguments, emergency-stop state, execution controls, the approved campaign limits, and the instrument method contract are checked before submission.

\textbf{Backend.} A service token preserves the room owner's project scope, verifies that the sample exists, and creates a \emph{submitted} measurement header before queuing.

\textbf{Device service.} The sidecar acquires an instrument claim and enforces the approved-method list, queue capacity, and instrument state; conflicts and unavailable resources become structured errors.

\textbf{Submission check.} Immediate status polling exposes launch failures, while device ownership, safety interlocks, approved methods, and operations already in progress remain under laboratory control.

\subsection{Campaign history and offline evaluation}

The experiment database retains raw files, measurements, analyses, notes, and versioned plans. MongoDB stores conversations, tool calls, plan state, and the per-well campaign ledger. Together they provide an ordered call--argument--result trace and support offline evaluation against configured or previously acquired files, as in Section~\ref{sec:campaign}. A watcher (a deterministic background process) polls the experiment database for completed measurements, copies the result files into the room, updates the ledger, and starts a new agent turn only when new evidence is available. Because each turn re-reads this saved state instead of relying on the earlier conversation, the loop can survive service restarts (Appendix~\ref{app:loop}).

\section{Physical Campaign and Archived-Data Evaluation}
\label{sec:campaign}

We assessed the architecture in two parts. During the physical campaign, the agent submitted measurements to the real instrument and interpreted results as they arrived. Later in the archived-data evaluation, the agent ran against a simulated instrument that returned files from the fully measured plate under controlled injection budgets; no reactions were rerun for any budget. The selected model reaction is the oxidative dimerization of \textit{p}-toluidine to 4,4$^\prime$-dimethylazobenzene, the benchmark introduced in Section~\ref{sec:intro} (Figure~\ref{fig:azo}). 

\begin{figure}[!htb]
\centering
\includegraphics[width=1\textwidth]{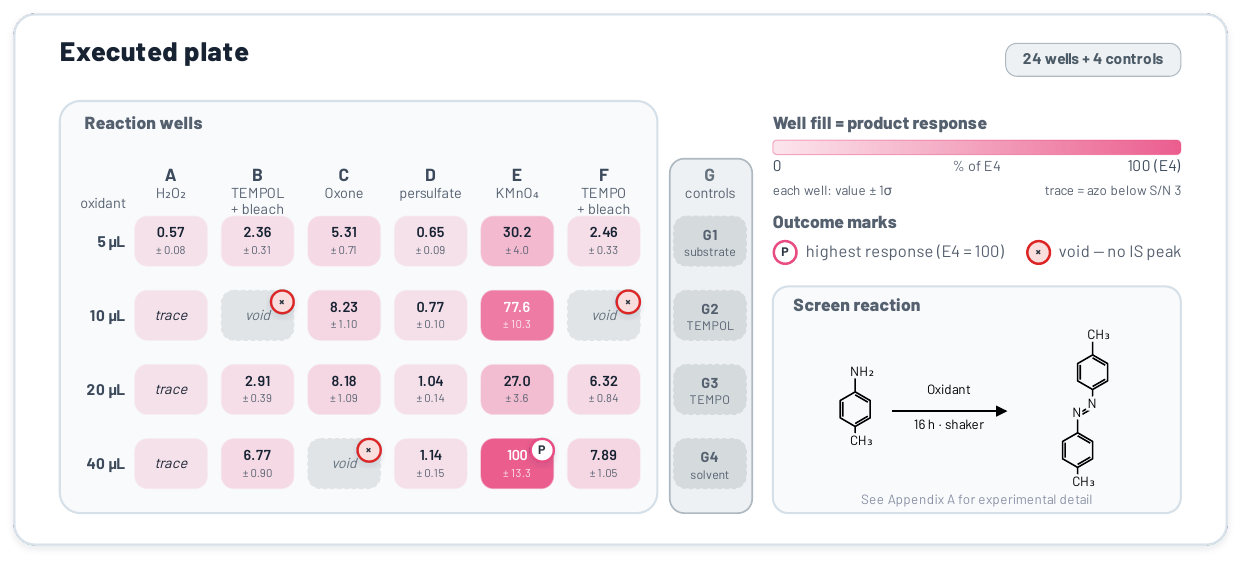}
\caption{Physical oxidant screen for \textit{p}-toluidine dimerization. Reactions were run in an Agilent 2.0~mL deep-well plate (5043-9302) at a total volume of 0.2~mL: 0.1~mL of 20~mM \textit{p}-toluidine stock in acetonitrile, plus oxidant and water. Full experimental details are given in Appendix~\ref{app:experiment}. The agent submits wells for LC--MS characterization, analyzes each result, and chooses the next injection; in this campaign, every well was measured at the operator's instruction.}
\label{fig:screen}
\end{figure}

The agent proposed the screen from the target transformation and the oxidants available in the laboratory, and the design was reviewed with experimental chemists: six oxidant systems (\ce{H2O2}, Oxone\textsuperscript{\textregistered}, ammonium persulfate, \ce{KMnO4}, and \ce{NaOCl}-mediated TEMPO and TEMPOL), each at 0.5, 1, 2, and 4 equivalents relative to \textit{p}-toluidine, plus four control wells, for 28 wells in total, all prepared in a single 96-well plate (Appendix~\ref{app:experiment}). The scientific objective was to find the well with the most product within a finite characterization budget. Figure~\ref{fig:screen} summarizes the plate layout and product response; experimental and evaluation procedures are provided in Appendix \ref{app:experiment}--\ref{app:simlab}, and prompts provided to the agents are recorded in Appendix \ref{app:prompts}.

\subsection{Physical oxidation screen}\label{sec:physical}

\begin{figure}[!htb]
\centering
\includegraphics[width=\textwidth]{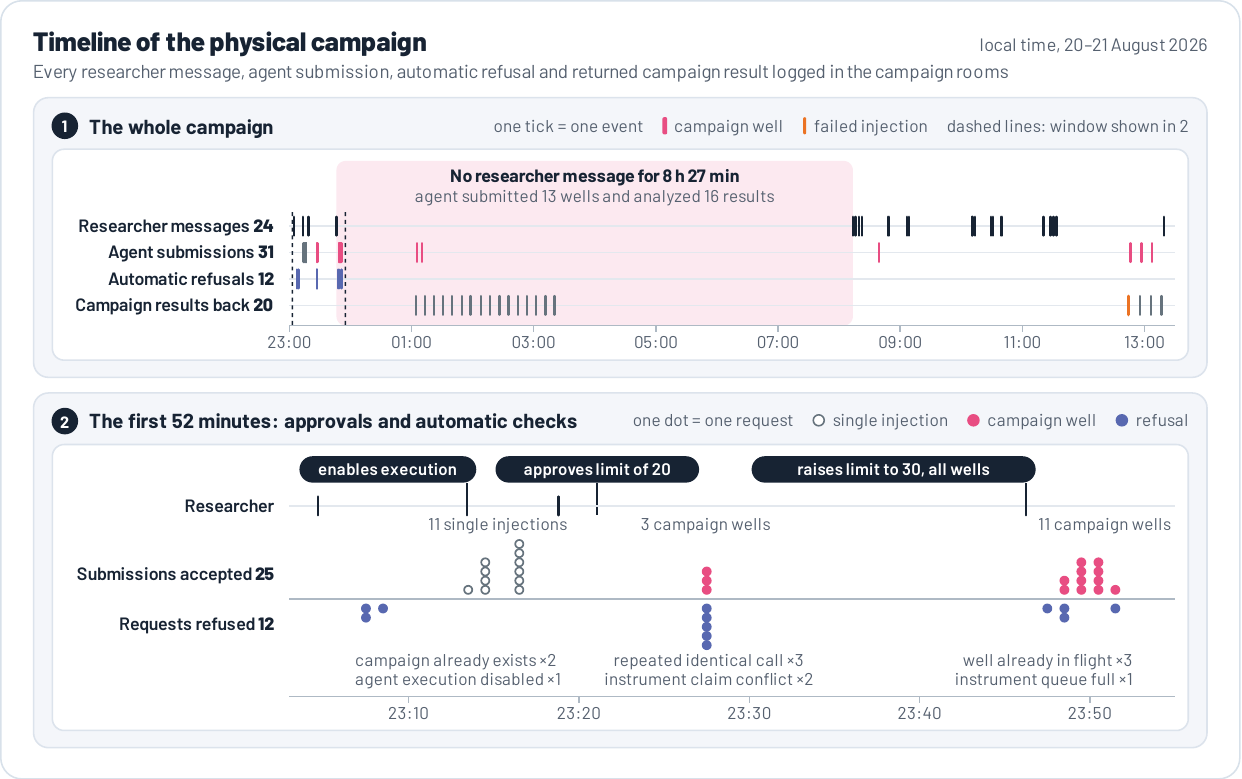}
\caption{Timeline of the physical campaign, drawn from logged events (local time). (1) The full campaign, showing researcher messages, agent submissions (11 pre-campaign single-well injections and 20 campaign injections), automatic refusals, and returned campaign results. The shaded interval marks the longest period without a researcher message (8~h~27~min), during which the agent submitted 13 wells and analyzed 16 results. (2) The first 52 minutes, showing the researcher's three control actions and all 12 automatic refusals listed in Table~\ref{tab:guards}.}
\label{fig:timeline}
\end{figure}

The physical LC--MS campaign (Figure~\ref{fig:timeline}) yielded usable data from 26 of 28 wells. Each injection corresponded to one LC--MS run of a single sample. All 28 wells were injected at least once. Twelve injections occurred outside the campaign and did not count toward its limit: 11 single injections before campaign approval and one solvent blank during method development. The campaign itself used 20 injections, within the updated limit of 30. It measured each of the 16 previously unmeasured wells once, injected B2 an additional time, and repeated B2, E2, and F2. The agent selected the measurement order and identified wells for repeat analysis. It flagged B2 and C4 as possible failed sample transfers; neither yielded usable data. It also flagged F2 as a low-quality draw, which was excluded from the ultraviolet (UV) quantification of the product (Figure~\ref{fig:screen}).
Two events illustrate the architecture's authority boundaries. First, the execution gate refused an agent request to submit an injection before the researcher enabled agent execution (Table~\ref{tab:guards}). Instrument access therefore remained conditional on explicit authorization. Second, approximately 20 minutes after approving the campaign, the researcher requested that every well be measured and increased the injection limit from 20 to 30. Because the interface did not support amendments to an approved campaign, the change was written directly to the campaign record, bypassing the formal approval workflow. The researcher had specified exhaustive well coverage; the agent retained discretion over measurement order and repeat selection, but not over which wells to leave unmeasured. This exposed a limitation of the architecture: campaign limits established through the approval workflow could subsequently be changed outside it.

\ce{KMnO4} produced a higher product response than every other oxidant, with E4 as the strongest well (Figure~\ref{fig:screen}). The best wells of Oxone\textsuperscript{\textregistered}, TEMPO/\ce{NaOCl}, and TEMPOL/\ce{NaOCl} produced 6.8--8.2\% of E4's response in the agent's internal-standard-normalized UV quantification and 7.4--8.2\% in an independent integration of the same chromatograms in the vendor software, Agilent OpenLab CDS (Appendix~\ref{app:experiment}, Table~\ref{tab:app-integration-comparison}). These differences were comparable to the estimated relative measurement uncertainty of approximately 13\%, so the three systems could not be reliably ranked against one another. Ammonium persulfate and \ce{H2O2} produced approximately 1\% or less of E4's response. Each condition was tested in a single well, without replicate reactions. The hypochlorite systems showed near-complete substrate conversion and far less product than \ce{KMnO4}, so conversion alone would have selected suboptimal conditions.

The product signal was assigned to 4,4$^\prime$-dimethylazobenzene. The agent identified the protonated molecular ion [M+H]$^+$ at $m/z$ 211, supported by chromatographic, spectral, isotope, and control-based evidence. Three organic chemists independently reviewed the assignment and judged the feature consistent with the proposed product. In the weakest conditions, isotope ratios inconsistent with a \ce{C14} ion revealed a co-eluting carbon-rich impurity; the agent therefore reported these product responses as upper bounds. An authentic 4,4$^\prime$-dimethylazobenzene standard, synthesized following Ruiz-Soriano \textit{et al.} \citep{Rui24}, reproduced the chromatographic and mass-spectral signatures of the assigned feature, confirming its identity. The product responses in Figure~\ref{fig:screen} come from an absorbance-based readout calibrated with this standard (Appendix~\ref{app:experiment}).

\textbf{Checks exercised and failures.} During the physical campaign, the agent made 465 tool calls in the main campaign room. The software checks of Section~\ref{sec:checks} verify that a request complies with the campaign limits, uses an approved method, and reaches the instrument only when it is available. They blocked or deferred 12 requests (Table~\ref{tab:guards}). Nine were stopped within the agent process, before leaving the laboratory server; the other three were rejected by the device service on the instrument PC. Two refusals occurred when the agent attempted to start a campaign that had already been created. The remaining ten occurred during individual well submissions. Each refusal returned an error explaining why the request could not proceed, and the agent adjusted its approach. When agent execution was disabled, its attempted injection submission was blocked, and it instead posted the request in the project room for the researcher to review and enable execution. When another session was using the instrument, submissions were refused and the agent resubmitted the wells individually. When the instrument queue was full, the agent deferred its request. No submission used an unapproved method, and the injection budget was not exceeded.

\begin{table}[H]
\centering
\small
\caption{Automatic software checks that refused or deferred agent requests during the physical campaign. Each row represents one type of check. \emph{Count} gives the number of requests refused or deferred, out of 465 tool calls made by the agent in the main campaign room. \emph{Enforced by} identifies where the check runs (Section~\ref{sec:checks}): checks in the agent process stop requests on the laboratory server before they are sent, while checks in the device service are enforced by the sidecar service on the instrument PC. All checks are applied by the software architecture, not by the agent or researcher. \emph{What it prevented} describes the action that would otherwise have occurred.}
\label{tab:guards}
\begin{tabular}{lcll}
\toprule
Check & Count & Enforced by & What it prevented \\
\midrule
Agent execution disabled & 1 & Agent process & a direct injection before the human enabled execution \\
Campaign already exists & 2 & Agent process & a second set of campaign limits on the same experiment \\
Instrument claim conflict & 2 & Device service & concurrent submissions racing for the instrument \\
Repeated identical call & 3 & Agent process & a runaway resubmission loop \\
Well already queued & 3 & Agent process & spending budget twice on a queued well \\
Instrument queue full & 1 & Device service & overrunning the instrument's 20-run queue \\
\bottomrule
\end{tabular}
\end{table}

Wells B2 and C4 ended the campaign without usable data. After the last campaign injection, the researcher asked for both wells to be queued again. The agent resubmitted B2, and the device service refused the request because the instrument was at that time being operated directly in the vendor software. No further injection of C4 was logged. This refusal occurred after the trace summarized in Table~\ref{tab:guards} had been exported, and it is not counted there.

\subsection{Archived-data evaluation}\label{sec:replay}

Because every well was measured in the physical campaign, that campaign does not show how well the agent chooses which wells to measure when its budget is limited. To test this choice, we reused the data from the fully measured plate. The same agent, analysis tools, watcher, and submission checks as in the physical campaign ran unchanged; only the instrument was replaced by a simulator, which returned the previously acquired LC–MS file for a well only once the agent had selected that well. No new reactions or measurements were made. We planned three simulated campaigns at each of nine injection budgets (1, 2, 4, 8, 12, 16, 20, 24, and 28 injections). Because larger budgets require many more agent calls and are therefore costly, we stopped after one campaign each at 20 and 28 injections, giving 23 campaigns in total. All 23 campaigns ran to completion. The best condition on the fully measured plate, 4 equivalents of \ce{KMnO4} in well E4, served as the reference: a campaign counts as successful if the agent identified E4 as the best well.

\begin{figure}[!htb]
\centering
\includegraphics[width=0.72\textwidth]{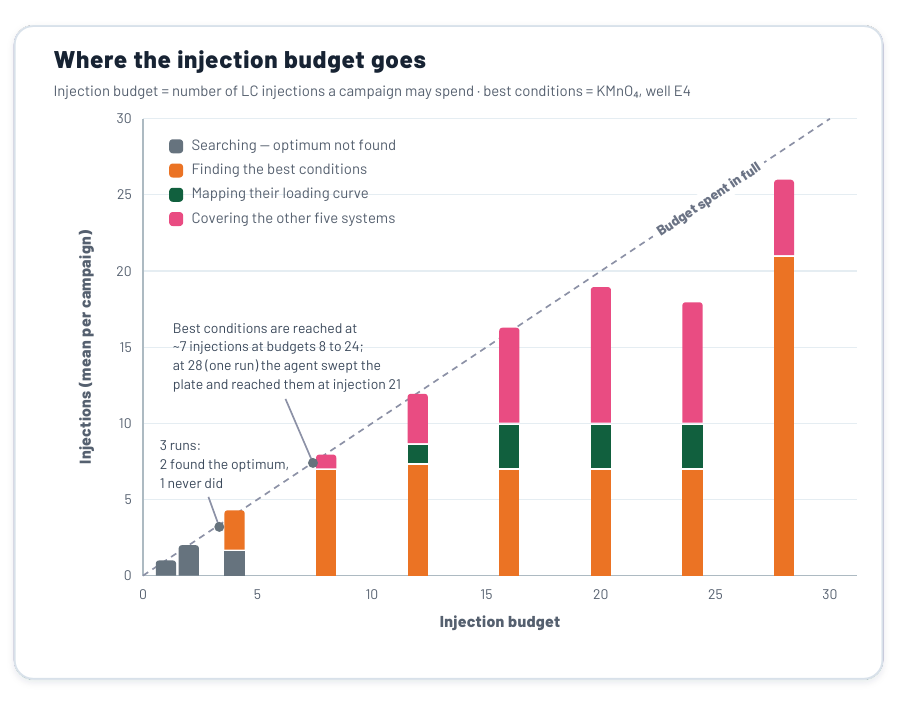}
\caption{Archived-data evaluation: the agent (Claude Opus 5, \texttt{anthropic:claude-opus-5}) runs against a simulated instrument that returns the previously acquired file for each well it selects, under different injection budgets. Three runs per budget, except one run each at budgets 20 and 28.}
\label{fig:budget}
\end{figure}

Figure~\ref{fig:budget} shows how the agent's performance depended on its injection budget. With budgets of 1 or 2 injections, no campaign identified the best condition. At 4 injections, two of the three campaigns succeeded; with budgets of 8 injections or more, every campaign identified the best condition. Because each budget was tested in at most three campaigns on a single plate, these results indicate a trend rather than precise success rates. Across all campaigns, the agent never reported a condition as best unless it had measured it.

Beyond identifying the best condition, additional injections expanded the information recovered. The agent typically first measured the best condition around its seventh injection. It then used the remaining budget to characterize how product response varied with \ce{KMnO4} loading and to measure the other five oxidant systems. With budgets of 12 injections or more, every campaign recovered the complete ranking of the six oxidant systems. Thus, once the budget was sufficient to identify the best condition, additional injections primarily improved the breadth of the characterization rather than the likelihood of finding the optimum.

For reference, we compared the agent's success with random well selection under the same budgets. A random selection of $b$ wells from the 28-well plate contains the best well with probability $b/28$: 14\% at 4 injections, 29\% at 8, and 43\% at 12. Under a random ordering of the plate, the best well would be measured at injection 14.5 on average, and 27 injections would be needed for a 95\% probability of measuring it. By comparison, the agent typically found the best condition around its seventh injection. Random selection provides a simple reference, not a fair baseline against other informed selection strategies. Fixed measurement schedules, greedy heuristics, and Bayesian optimization could also exploit the plate's structure. Comparing against these approaches, as well as benchmarking other language models, is outside the scope of this demonstration and left for future work. We evaluated only one language model, Claude Opus 5; these results therefore characterize what this agent achieved on this particular plate rather than establish a general measure of agent performance (Section~\ref{sec:limitations}).

Adaptive selection also incurs a reasoning cost. Unlike random or fixed selection rules, which can choose the next well almost immediately, the agent must inspect newly acquired data and reason about its next measurement. Across campaigns with budgets of 8--24 injections, logged reasoning time totalled 10--24 minutes per campaign. The simulated instrument assumed 15 minutes of acquisition per injection, making even an 8-injection campaign approximately two hours long. Reasoning therefore added a modest overhead relative to acquisition time. This overhead was offset by the earlier identification of the best condition: finding it around injection 7 rather than the average injection 14.5 under random ordering saves approximately 7.5 injections, or nearly two hours of instrument time. For instruments with substantially shorter acquisition times, however, reasoning would account for a larger fraction of campaign time and could limit throughput.

The budget also influenced how the agent searched. With a budget of 28 injections, sufficient to measure every well, the agent worked through the plate sequentially and did not measure the best condition until its 21st injection. Under tighter budgets, it instead prioritized wells it expected to be informative, typically finding the best condition around its seventh injection. In this campaign, a generous budget therefore led to exhaustive coverage rather than rapid identification of the best condition. This suggests that the agent's search strategy responds to the resources available: it prioritizes measurements when resources are scarce but defaults to broader coverage when they are not. The injection budget consequently serves not only as an execution limit but also as a determinant of search behaviour. It should be specified deliberately by a human expert as part of the approved campaign limits, with the campaign objective making clear whether the priority is rapid identification of the best condition or comprehensive characterization of the plate. Because the 28-injection budget was tested in only one campaign, this behavioural observation requires confirmation with additional replicates.
\section{Scope and Limitations}
\label{sec:limitations}

\textbf{What the agent controlled.} The demonstrated autonomy is at the \dmta{A}nalyze--\dmta{D}esign interface, not the full DMTA cycle. Humans designed and prepared the reaction plate, approved the campaign limits, enabled agent execution, and extended the physical campaign to every well. The agent submitted and ordered measurements, analyzed results, flagged failed transfers and low-quality draws, and ranked conditions. It did not prepare or synthesize any reaction.

\textbf{Breadth.} This study covers one reaction system, one instrument workflow, and one experimental campaign; its contribution is a bounded agent--laboratory workflow rather than a general solution to chemical discovery. Testing the architecture more broadly will require campaigns that differ from this one. On the same LC--MS instrument, other chemistry experiments, such as reaction kinetics studies, in which the agent must decide when to sample rather than which well, and larger high-throughput screens would test whether its analysis and measurement selection carry over to different scientific questions. Other analytical instruments, such as gas chromatography--mass spectrometry or nuclear magnetic resonance spectroscopy, would test whether the same authorization and validation checks hold when the data and instrument services change. Finally, integrating the architecture with other SDL platforms, including automated synthesis, and coupling it with other agents, such as La Agente \'Optima, which constructs and supervises Bayesian optimization campaigns \citep{Mul26b}, would test whether the same safeguards hold when the agent's decisions trigger new reactions rather than only new measurements.

\textbf{Selection quality and cost.} Because the archived-data evaluation does not compare against non-agentic selection policies, it establishes feasibility on this plate rather than a comparative efficiency advantage or a cost--benefit case for agentic selection. The eight-injection result holds for this plate, model, prompts, and evaluation setup. A fair comparison would run random sampling over fixed seeds, a fixed or greedy schedule, and a suitably tuned Bayesian optimizer under the same information, budget, and scoring rules, and would score finding the best well, finding the best oxidant and loading, and recovering the full ranking separately.

\textbf{Safety and intervention.} The checks in Table~\ref{tab:guards} were exercised in physical operation; the emergency stop and the \emph{Stop} control were not. The emergency stop refuses further submissions and holds campaign wake-ups until a human releases it (Appendix~\ref{app:ui-agent}), but we have not tested a campaign in which a predefined, undesirable condition triggers human review, pauses execution, and resumes or aborts on the human's decision. Amending an approved campaign also had no path through the interface: the budget change in this campaign was written directly to the campaign record, outside the approval step. Finally, the checks in this work are operational rather than chemical: they control which actions reach the instrument, not whether a proposed experiment is hazardous. Coupling the architecture with a chemical-safety agent such as El Agente Seguro \citep{Kang26} could add a hazard assessment before campaign limits are approved.

\textbf{Next steps.} The next extension is to use the same validated workflow for preparation: the agent would select the next condition, an operator or approved automation would prepare it, and its characterization result would inform the next experiment. A second reaction system and instrument, failed-well repeats, a tested intervention trigger, and a single event log covering sample preparation, instrument use, agent choices, and human interventions remain future validation steps.

\section{Conclusion}
\label{sec:conclusion}

Autonomy in self-driving laboratories should be achieved by defining where an agent may act, not by giving it unrestricted control of hardware. Our architecture grants the agent limited autonomy at the Analyze--Design interface of the DMTA cycle and makes its interfaces explicit: the agent may interpret results and select measurements within researcher-approved campaign limits, but it may not reach the instrument directly or change those limits (Section~\ref{sec:architecture}). The SDL network connects the instrument, the experiment database, the agent service, and the researchers' shared project room, so that every agent request is validated and logged by laboratory services, and execution begins only after a researcher approves the campaign and enables it. In a physical LC--MS oxidant screen, the agent ordered the measurements, interpreted MS and DAD evidence, flagged failed sample transfers, and identified the best-performing oxidant and loading, a ranking reproduced by independent peak integration (Section~\ref{sec:physical}). In replays of the archived data under tight injection budgets, the agent prioritized informative wells and, once the budget was sufficient, reached the best condition earlier than random ordering would on average (Section~\ref{sec:replay}). Together, these results demonstrate that an agent can direct measurement selection and interpret experimental data on a physical instrument without having direct authority over its operation. Extending this approach to sample preparation would test whether the same safeguards can support agent-directed decisions across a larger portion of the DMTA cycle.

\section*{Acknowledgments}
This research is part of the University of Toronto’s Acceleration Consortium, which is funded by the CFREF-2022-00042 Canada First Research Excellence Fund. This work was also supported by the Defense Advanced Research Projects Agency (DARPA) under Agreement No. HR0011262E022 and the AI2050 program of Schmidt
Sciences. A.A.-G. also acknowledges the generous support of Natural Resources Canada and the Canada 150 Research Chairs program. A.A.-G. thanks Anders G. Frøseth for his generous support.

\clearpage

{
\small
\bibliography{references}

\begin{thebibliography}{49}
\providecommand{\natexlab}[1]{#1}
\providecommand{\url}[1]{\texttt{#1}}
\expandafter\ifx\csname urlstyle\endcsname\relax
  \providecommand{\doi}[1]{doi: #1}\else
  \providecommand{\doi}{doi: \begingroup \urlstyle{rm}\Url}\fi

\bibitem[H{\"a}se et~al.(2019)H{\"a}se, Roch, and Aspuru-Guzik]{Has19}
F.~H{\"a}se, L.~Roch, and A.~Aspuru-Guzik.
\newblock Next-generation experimentation with self-driving laboratories.
\newblock \emph{Trends in Chemistry}, 1:\penalty0 282--291, 2019.
\newblock \doi{10.1016/j.trechm.2019.02.007}.

\bibitem[Tom et~al.(2024)Tom, Schmid, Baird, Cao, Darvish, Hao, Lo, Pablo-García, Rajaonson, Skreta, Yoshikawa, Corapi, Akkoc, Strieth-Kalthoff, Seifrid, and Aspuru-Guzik]{Tom24}
Gary Tom, Stefan~P. Schmid, Sterling~G. Baird, Yang Cao, Kourosh Darvish, Han Hao, Stanley Lo, Sergio Pablo-García, Ella~M. Rajaonson, Marta Skreta, Naruki Yoshikawa, Samantha Corapi, Gun~Deniz Akkoc, Felix Strieth-Kalthoff, Martin Seifrid, and Alán Aspuru-Guzik.
\newblock Self-driving laboratories for chemistry and materials science.
\newblock \emph{Chemical Reviews}, 124:\penalty0 9633--9732, 2024.
\newblock \doi{10.1021/acs.chemrev.4c00055}.

\bibitem[Aspuru-Guzik and Bernales(2026)]{Asp26}
A.~Aspuru-Guzik and V.~Bernales.
\newblock Reaction: Building trust infrastructure for agentic experimental science beyond fragile eloquence.
\newblock \emph{Chem}, 12\penalty0 (8):\penalty0 103163, 2026.
\newblock \doi{10.1016/j.chempr.2026.103163}.

\bibitem[Scheurer and Reuter(2025)]{Sch25}
Christoph Scheurer and Karsten Reuter.
\newblock Role of the human-in-the-loop in emerging self-driving laboratories for heterogeneous catalysis.
\newblock \emph{Nature Catalysis}, 8\penalty0 (1):\penalty0 13--19, 2025.
\newblock \doi{10.1038/s41929-024-01275-5}.

\bibitem[Leong et~al.(2025)Leong, Griesbach, Zhang, Darvish, Zhao, Mandal, Zou, Hao, Bernales, and Aspuru-Guzik]{Leo25}
Shi~Xuan Leong, Caleb~E. Griesbach, Rui Zhang, Kourosh Darvish, Yuchi Zhao, Abhijoy Mandal, Yunheng Zou, Han Hao, Varinia Bernales, and Alán Aspuru-Guzik.
\newblock Steering towards safe self-driving laboratories.
\newblock \emph{Nature Reviews Chemistry}, pages 707--722, 2025.
\newblock \doi{10.1038/s41570-025-00747-x}.

\bibitem[Kang et~al.(2026)Kang, Leong, Park, Bai, Papidocha, Zhao, Zhang, Gimeno, Lo, Isbrandt, Hao, Cao, Katzenburg, Du, Aspuru-Guzik, and Bernales]{Kang26}
Yeonghun Kang, Shi~Xuan Leong, Hyun~Suk Park, Jiaru Bai, Sven Papidocha, Yuchi Zhao, Rui Zhang, Magali Gimeno, Stanley Lo, Eric~S. Isbrandt, Han Hao, Yang Cao, Felix Katzenburg, Fangshi Du, Alán Aspuru-Guzik, and Varinia Bernales.
\newblock El {Agente Seguro}: An agent for chemical safety.
\newblock \emph{ChemRxiv}, 2026.
\newblock \doi{10.26434/chemrxiv.15004378/v2}.
\newblock June 10.

\bibitem[Lee et~al.(2026)Lee, Yoo, Jang, Park, Park, and Han]{Lee26}
Heeseung Lee, Hyuk~Jun Yoo, Hye~Su Jang, Byeongho Park, Yang~Jeong Park, and Sang~Soo Han.
\newblock Toward self-driving laboratory 2.0 for chemistry and materials discovery.
\newblock \emph{Materials Horizons}, 2026.
\newblock \doi{10.1039/d5mh01984b}.

\bibitem[Xie et~al.(2026)Xie, Luo, Ye, Chen, Zhu, and Jiang]{Xie26}
Zikai Xie, Man Luo, Zixin Ye, Linjiang Chen, Zhuoying Zhu, and Jun Jiang.
\newblock Autonomous chemistry and materials innovation driven by scientific agents.
\newblock \emph{JACS Au}, pages 2659--2669, 2026.
\newblock \doi{10.1021/jacsau.6c00213}.

\bibitem[Chen et~al.(2026)Chen, Wang, Yin, Jiang, and Zhang]{Che26}
Xuanzhou Chen, Audrey Wang, Stanley Yin, Hanyang Jiang, and Dong Zhang.
\newblock Agentic {AI} for self-driving laboratories in soft matter: Taxonomy, benchmarks, and open challenges.
\newblock \emph{arXiv preprint arXiv:2601.17920}, 2026.
\newblock \doi{10.48550/arXiv.2601.17920}.

\bibitem[Zhang et~al.(2026)Zhang, Que, Chang, Zhang, Wei, and Zhu]{Zha26b}
Zihan Zhang, Haohui Que, Junhan Chang, Xin Zhang, Hao Wei, and Tong Zhu.
\newblock Safe-{SDL}: Establishing safety boundaries and control mechanisms for {AI}-driven self-driving laboratories.
\newblock \emph{arXiv preprint arXiv:2602.15061}, 2026.
\newblock \doi{10.48550/arXiv.2602.15061}.

\bibitem[Strickland et~al.(2026)Strickland, Vijeta, Moores, Bodek, Nenchev, Whitehead, Phillips, Tassenberg, Conduit, and Pellegrini]{Str26}
Joel Strickland, Arjun Vijeta, Chris Moores, Oliwia Bodek, Bogdan Nenchev, Thomas Whitehead, Charles Phillips, Karl Tassenberg, Gareth Conduit, and Ben Pellegrini.
\newblock Talk freely, execute strictly: Schema-gated agentic {AI} for flexible and reproducible scientific workflows.
\newblock \emph{arXiv preprint arXiv:2603.06394}, 2026.
\newblock \doi{10.48550/arXiv.2603.06394}.

\bibitem[Adamidis et~al.(2026)Adamidis, Katrisioti, Tzitzikas, and Stratakis]{Ada26}
Marios Adamidis, Danae Katrisioti, Yannis Tzitzikas, and Emmanuel Stratakis.
\newblock It's not the language model, it's the tool: Deterministic mediation for scientific workflows.
\newblock \emph{arXiv preprint arXiv:2605.13245}, 2026.
\newblock \doi{10.48550/arXiv.2605.13245}.

\bibitem[Yoshikawa et~al.(2023)Yoshikawa, Skreta, Darvish, Arellano-Rubach, Ji, Bjørn~Kristensen, Li, Zhao, Xu, Kuramshin, Aspuru-Guzik, Shkurti, and Garg]{Yos23}
Naruki Yoshikawa, Marta Skreta, Kourosh Darvish, Sebastian Arellano-Rubach, Zhi Ji, Lasse Bjørn~Kristensen, Andrew~Zou Li, Yuchi Zhao, Haoping Xu, Artur Kuramshin, Alán Aspuru-Guzik, Florian Shkurti, and Animesh Garg.
\newblock Large language models for chemistry robotics.
\newblock \emph{Autonomous Robots}, pages 1057--1086, 2023.
\newblock \doi{10.1007/s10514-023-10136-2}.

\bibitem[Pagel et~al.(2026)Pagel, Jirasek, and Cronin]{Pag26}
S.~Pagel, M.~Jirasek, and L.~Cronin.
\newblock Verification and execution of the scientific literature via chemputation augmented by large language models.
\newblock \emph{Communications Chemistry}, 2026.
\newblock \doi{10.1038/s42004-026-01993-w}.

\bibitem[Eltaboni et~al.(2022)Eltaboni, Bader, El-Kailany, Elsharif, and Ahmida]{eltaboni2022azo}
F.~Eltaboni, N.~Bader, R.~El-Kailany, N.~Elsharif, and A.~Ahmida.
\newblock Chemistry and applications of azo dyes: A comprehensive review.
\newblock \emph{J. Chem. Rev.}, 4\penalty0 (4):\penalty0 313--330, 2022.
\newblock \doi{10.22034/jcr.2022.349827.1177}.

\bibitem[Georgi{\'a}des et~al.(2015)Georgi{\'a}des, {\"O}tv{\"o}s, and F{\"u}l{\"o}p]{Geo15}
{\'A}d{\'a}m Georgi{\'a}des, S{\'a}ndor~B. {\"O}tv{\"o}s, and Ferenc F{\"u}l{\"o}p.
\newblock Exploring new parameter spaces for the oxidative homocoupling of aniline derivatives: Sustainable synthesis of azobenzenes in a flow system.
\newblock \emph{ACS Sustainable Chemistry \& Engineering}, 3\penalty0 (12):\penalty0 3388--3397, 2015.
\newblock \doi{10.1021/acssuschemeng.5b01096}.

\bibitem[Wu et~al.(2008)Wu, Ondruschka, Cravotto, Garella, and Asgari]{Wu08}
Zhi-Lin Wu, Bernd Ondruschka, Giancarlo Cravotto, Davide Garella, and Jila Asgari.
\newblock Oxidation of primary aromatic amines under irradiation with ultrasound and/or microwaves.
\newblock \emph{Synthetic Communications}, 38\penalty0 (15):\penalty0 2619--2624, 2008.
\newblock \doi{10.1080/00397910802219387}.

\bibitem[Geysen et~al.(1984)Geysen, Meloen, and Barteling]{Gey84}
H.~M. Geysen, R.~H. Meloen, and S.~J. Barteling.
\newblock Use of peptide synthesis to probe viral antigens for epitopes to a resolution of a single amino acid.
\newblock \emph{Proceedings of the National Academy of Sciences of the United States of America}, 81\penalty0 (13):\penalty0 3998--4002, 1984.
\newblock \doi{10.1073/pnas.81.13.3998}.

\bibitem[Devore and Jenkins(2014)]{Dev14}
D.~D. Devore and R.~M. Jenkins.
\newblock Impact of high throughput experimentation on homogeneous catalysis research.
\newblock \emph{Comments on Inorganic Chemistry}, 34\penalty0 (1--2):\penalty0 17--41, 2014.
\newblock \doi{10.1080/02603594.2014.947027}.

\bibitem[Fitzpatrick et~al.(2016)Fitzpatrick, Battilocchio, and Ley]{Fit15}
Daniel~E. Fitzpatrick, Claudio Battilocchio, and Steven~V. Ley.
\newblock A novel internet-based reaction monitoring, control and autonomous self-optimization platform for chemical synthesis.
\newblock \emph{Organic Process Research \& Development}, 20\penalty0 (2):\penalty0 386--394, 2016.
\newblock \doi{10.1021/acs.oprd.5b00313}.

\bibitem[Steiner et~al.(2019)Steiner, Wolf, Glatzel, Andreou, Granda, Keenan, Hinkley, Aragon-Camarasa, Kitson, Angelone, and Cronin]{Ste19}
Sebastian Steiner, Jakob Wolf, Stefan Glatzel, Anna Andreou, Jarosław~M. Granda, Graham Keenan, Trevor Hinkley, Gerardo Aragon-Camarasa, Philip~J. Kitson, Davide Angelone, and Leroy Cronin.
\newblock Organic synthesis in a modular robotic system driven by a chemical programming language.
\newblock \emph{Science}, 2019.
\newblock \doi{10.1126/science.aav2211}.

\bibitem[Mehr et~al.(2020)Mehr, Craven, Leonov, Keenan, and Cronin]{Meh20}
S.~Hessam~M. Mehr, Matthew Craven, Artem~I. Leonov, Graham Keenan, and Leroy Cronin.
\newblock A universal system for digitization and automatic execution of the chemical synthesis literature.
\newblock \emph{Science}, pages 101--108, 2020.
\newblock \doi{10.1126/science.abc2986}.

\bibitem[Rohrbach et~al.(2022)Rohrbach, Šiaučiulis, Chisholm, Pirvan, Saleeb, Mehr, Trushina, Leonov, Keenan, Khan, Hammer, and Cronin]{Roh22}
Simon Rohrbach, Mindaugas Šiaučiulis, Greig Chisholm, Petrisor-Alin Pirvan, Michael Saleeb, S.~Hessam~M. Mehr, Ekaterina Trushina, Artem~I. Leonov, Graham Keenan, Aamir Khan, Alexander Hammer, and Leroy Cronin.
\newblock Digitization and validation of a chemical synthesis literature database in the chempu.
\newblock \emph{Science}, 377\penalty0 (6602):\penalty0 172--180, 2022.
\newblock \doi{10.1126/science.abo0058}.

\bibitem[Rauschen et~al.(2024)Rauschen, Guy, Hein, and Cronin]{Rau24}
Robert Rauschen, Mason Guy, J.~Hein, and L.~Cronin.
\newblock Universal chemical programming language for robotic synthesis repeatability.
\newblock \emph{Nature Synthesis}, 3:\penalty0 488--496, 2024.
\newblock \doi{10.1038/s44160-023-00473-6}.

\bibitem[{\v{S}}iau{\v{c}}iulis et~al.(2024){\v{S}}iau{\v{c}}iulis, Knittl-Frank, Mehr, Clarke, and Cronin]{Sia24}
Mindaugas {\v{S}}iau{\v{c}}iulis, Christian Knittl-Frank, S.~H.~M. Mehr, Emma Clarke, and L.~Cronin.
\newblock Reaction blueprints and logical control flow for parallelized chiral synthesis in the chemputer.
\newblock \emph{Nature Communications}, 15:\penalty0 10261, 2024.
\newblock \doi{10.1038/s41467-024-54238-6}.

\bibitem[Fakhruldeen et~al.(2022)Fakhruldeen, Pizzuto, Glowacki, and Cooper]{Fak22}
Hatem Fakhruldeen, Gabriella Pizzuto, J.~Glowacki, and Andrew~I. Cooper.
\newblock {ARChemist}: Autonomous robotic chemistry system architecture.
\newblock In \emph{2022 IEEE International Conference on Robotics and Automation}, pages 6013--6019, 2022.
\newblock \doi{10.1109/ICRA46639.2022.9811996}.

\bibitem[Christensen et~al.(2021{\natexlab{a}})Christensen, Yunker, Shiri, Zepel, Prieto, Grunert, Bork, and Hein]{Chr21}
Melodie Christensen, Lars P.~E. Yunker, Parisa Shiri, Tara Zepel, Paloma~L. Prieto, Shad Grunert, Finn Bork, and Jason~E. Hein.
\newblock Automation isn't automatic.
\newblock \emph{Chemical Science}, 12:\penalty0 15473--15490, 2021{\natexlab{a}}.
\newblock \doi{10.1039/d1sc04588a}.

\bibitem[Roch et~al.(2020)Roch, Häse, Kreisbeck, Tamayo-Mendoza, Yunker, Hein, and Aspuru-Guzik]{Roc20}
Loïc~M. Roch, Florian Häse, Christoph Kreisbeck, Teresa Tamayo-Mendoza, Lars P.~E. Yunker, Jason~E. Hein, and Alán Aspuru-Guzik.
\newblock {ChemOS}: An orchestration software to democratize autonomous discovery.
\newblock \emph{PLoS ONE}, 2020.
\newblock \doi{10.1371/journal.pone.0229862}.

\bibitem[Sim et~al.(2024)Sim, Vakili, Strieth-Kalthoff, Hao, Hickman, Miret, Pablo-García, and Aspuru-Guzik]{Sim24}
Malcolm Sim, Mohammad~Ghazi Vakili, Felix Strieth-Kalthoff, Han Hao, Riley~J. Hickman, Santiago Miret, Sergio Pablo-García, and Alán Aspuru-Guzik.
\newblock {ChemOS} 2.0: An orchestration architecture for chemical self-driving laboratories.
\newblock \emph{Matter}, 2024.
\newblock \doi{10.1016/j.matt.2024.04.022}.

\bibitem[Canty et~al.(2023)Canty, Koscher, McDonald, and Jensen]{Can23}
Richard~B. Canty, Brent~A. Koscher, Matthew~A. McDonald, and Klavs~F. Jensen.
\newblock Integrating autonomy into automated research platforms.
\newblock \emph{Digital Discovery}, 2:\penalty0 1259--1268, 2023.
\newblock \doi{10.1039/d3dd00135k}.

\bibitem[Pendleton et~al.(2019)Pendleton, Cattabriga, Li, Najeeb, Friedler, Norquist, Chan, and Schrier]{Pen19}
Ian~M. Pendleton, Gary Cattabriga, Zhi Li, Mansoor~Ani Najeeb, Sorelle~A. Friedler, Alexander~J. Norquist, Emory~M. Chan, and Joshua Schrier.
\newblock {ESCALATE}: Experiment specification, capture and laboratory automation technology: A software pipeline for automated chemical experimentation and data management.
\newblock \emph{MRS Communications}, pages 846--859, 2019.
\newblock \doi{10.1557/MRC.2019.72}.

\bibitem[Christensen et~al.(2021{\natexlab{b}})Christensen, Yunker, Adedeji, Häse, Roch, Gensch, dos Passos~Gomes, Zepel, Sigman, Aspuru-Guzik, and Hein]{Chr20}
Melodie Christensen, Lars P.~E. Yunker, Folarin Adedeji, Florian Häse, Loïc~M. Roch, Tobias Gensch, Gabriel dos Passos~Gomes, Tara Zepel, Matthew~S. Sigman, Alán Aspuru-Guzik, and Jason~E. Hein.
\newblock Data-science driven autonomous process optimization.
\newblock \emph{Communications Chemistry}, 2021{\natexlab{b}}.
\newblock \doi{10.1038/s42004-021-00550-x}.

\bibitem[Angello et~al.(2022)Angello, Rathore, Beker, Wołos, Jira, Roszak, Wu, Schroeder, Aspuru-Guzik, Grzybowski, and Burke]{Ang22}
Nicholas~H. Angello, Vandana Rathore, Wiktor Beker, Agnieszka Wołos, Edward~R. Jira, Rafał Roszak, Tony~C. Wu, Charles~M. Schroeder, Alán Aspuru-Guzik, Bartosz~A. Grzybowski, and Martin~D. Burke.
\newblock Closed-loop optimization of general reaction conditions for heteroaryl suzuki--miyaura coupling.
\newblock \emph{Science}, pages 399--405, 2022.
\newblock \doi{10.1126/science.adc8743}.

\bibitem[Leonov et~al.(2024)Leonov, Hammer, Lach, Mehr, Caramelli, Angelone, Khan, O'Sullivan, Craven, Wilbraham, and Cronin]{Leo24}
Artem~I. Leonov, Alexander J.~S. Hammer, Slawomir Lach, S.~Hessam~M. Mehr, Dario Caramelli, Davide Angelone, Aamir Khan, Steven O'Sullivan, Matthew Craven, Liam Wilbraham, and Leroy Cronin.
\newblock An integrated self-optimizing programmable chemical synthesis and reaction engine.
\newblock \emph{Nature Communications}, 2024.
\newblock \doi{10.1038/s41467-024-45444-3}.

\bibitem[Strieth-Kalthoff et~al.(2024)Strieth-Kalthoff, Hao, Rathore, Derasp, Gaudin, Angello, Seifrid, Trushina, Guy, Liu, Tang, Mamada, Wang, Tsagaantsooj, Lavigne, Pollice, Wu, Hotta, Bodo, Li, Haddadnia, Wołos, Roszak, Ser, Bozal-Ginesta, Hickman, Vestfrid, Aguilar-Granda, Klimareva, Sigerson, Hou, Gahler, Lach, Warzybok, Borodin, Rohrbach, Sanchez-Lengeling, Adachi, Grzybowski, Cronin, Hein, Burke, and Aspuru-Guzik]{Str24}
Felix Strieth-Kalthoff, Han Hao, Vandana Rathore, Joshua Derasp, Théophile Gaudin, Nicholas~H. Angello, Martin Seifrid, Ekaterina Trushina, Mason Guy, Junliang Liu, Xun Tang, Masashi Mamada, Wesley Wang, Tuul Tsagaantsooj, Cyrille Lavigne, Robert Pollice, Tony~C. Wu, Kazuhiro Hotta, Leticia Bodo, Shangyu Li, Mohammad Haddadnia, Agnieszka Wołos, Rafał Roszak, Cher~Tian Ser, Carlota Bozal-Ginesta, Riley~J. Hickman, Jenya Vestfrid, Andrés Aguilar-Granda, Elena~L. Klimareva, Ralph~C. Sigerson, Wenduan Hou, Daniel Gahler, Slawomir Lach, Adrian Warzybok, Oleg Borodin, Simon Rohrbach, Benjamin Sanchez-Lengeling, Chihaya Adachi, Bartosz~A. Grzybowski, Leroy Cronin, Jason~E. Hein, Martin~D. Burke, and Alán Aspuru-Guzik.
\newblock Delocalized, asynchronous, closed-loop discovery of organic laser emitters.
\newblock \emph{Science}, 2024.
\newblock \doi{10.1126/science.adk9227}.

\bibitem[Bai et~al.(2022)Bai, Cao, Mosbach, Akroyd, Lapkin, and Kraft]{Bai22}
Jiaru Bai, Liwei Cao, S.~Mosbach, J.~Akroyd, A.~Lapkin, and Markus Kraft.
\newblock From platform to knowledge graph: Evolution of laboratory automation.
\newblock \emph{JACS Au}, 2:\penalty0 292--309, 2022.
\newblock \doi{10.1021/jacsau.1c00438}.

\bibitem[Bai et~al.(2024)Bai, Mosbach, Taylor, Karan, Lee, Rihm, Akroyd, Lapkin, and Kraft]{Bai24}
Jiaru Bai, Sebastian Mosbach, Connor~J. Taylor, Dogancan Karan, Kok~Foong Lee, Simon~D. Rihm, Jethro Akroyd, Alexei~A. Lapkin, and Markus Kraft.
\newblock A dynamic knowledge graph approach to distributed self-driving laboratories.
\newblock \emph{Nature Communications}, 2024.
\newblock \doi{10.1038/s41467-023-44599-9}.

\bibitem[Boiko et~al.(2023)Boiko, MacKnight, Kline, and Gomes]{Boi23}
Daniil~A. Boiko, Robert MacKnight, Ben Kline, and Gabe Gomes.
\newblock Autonomous chemical research with large language models.
\newblock \emph{Nature}, pages 570--578, 2023.
\newblock \doi{10.1038/s41586-023-06792-0}.

\bibitem[Bran et~al.(2024)Bran, Cox, Schilter, Baldassari, White, and Schwaller]{Bra23}
Andres~M. Bran, Sam Cox, Oliver Schilter, Carlo Baldassari, Andrew~D. White, and Philippe Schwaller.
\newblock Augmenting large language models with chemistry tools.
\newblock \emph{Nature Machine Intelligence}, pages 525--535, 2024.
\newblock \doi{10.1038/s42256-024-00832-8}.

\bibitem[Ruan et~al.(2024)Ruan, Lu, Xu, He, Chen, Zhang, Xuan, Pan, Fang, Gao, Shen, Ye, Zhang, and Mo]{Rua24}
Yixiang Ruan, Chenyin Lu, Ning Xu, Yuchen He, Yixin Chen, Jian Zhang, Jun Xuan, Jianzhang Pan, Qun Fang, Hanyu Gao, Xiaodong Shen, Ning Ye, Qiang Zhang, and Yiming Mo.
\newblock An automatic end-to-end chemical synthesis development platform powered by large language models.
\newblock \emph{Nature Communications}, 2024.
\newblock \doi{10.1038/s41467-024-54457-x}.

\bibitem[Song et~al.(2025)Song, Luo, Zhang, Chen, Huang, Cao, Zhu, Liu, Zhang, Zou, Zhang, Zhang, Shang, Fu, Jiang, and Luo]{Son25}
Tao Song, Man Luo, Xiaolong Zhang, Linjiang Chen, Yan Huang, Jiaqi Cao, Qing Zhu, Daobin Liu, Baicheng Zhang, Gang Zou, Guoqing Zhang, Fei Zhang, Weiwei Shang, Yao Fu, Jun Jiang, and Yi~Luo.
\newblock A multiagent-driven robotic {AI} chemist enabling autonomous chemical research on demand.
\newblock \emph{Journal of the American Chemical Society}, 2025.
\newblock \doi{10.1021/jacs.4c17738}.

\bibitem[Müller et~al.(2026)Müller, Bai, Gottstein, Mandal, Nazeri, Savino, Fang, Das, Carrillo, Kang, Pérez-Sánchez, Pilon, Fitzner, Noël, Gu, Bernales, and Aspuru-Guzik]{Mul26b}
Marcel Müller, Jiaru Bai, Willi Gottstein, Abhijoy Mandal, Mohammad Nazeri, Elia Savino, Yanlin Fang, Sujoy Das, Sergio Pablo~García Carrillo, Yeonghun Kang, Juan~B. Pérez-Sánchez, Simone Pilon, Martin Fitzner, Timothy Noël, Frank Gu, Varinia Bernales, and Alán Aspuru-Guzik.
\newblock La agente {\'o}ptima: Towards agentic self-driving laboratories, 2026.
\newblock URL \url{https://arxiv.org/abs/2609.04564}.

\bibitem[Cai et~al.(2026)Cai, Zhao, Yue, Liu, Liu, Cui, Zhou, Tian, Zhang, Yang, Zhu, and Mo]{Cai26}
Wendi Cai, Boxuan Zhao, Yansong Yue, Changlin Liu, Chengchun Liu, Ying Cui, Zihao Zhou, Xiaohui Tian, Zhongchao Zhang, Zhen Yang, Jie Zhu, and Fanyang Mo.
\newblock Agentic preparative thin-layer chromatography system for autonomous purification.
\newblock \emph{Journal of the American Chemical Society}, pages 43269--43279, 2026.
\newblock \doi{10.1021/jacs.6c10800}.

\bibitem[Fehlis et~al.(2025)Fehlis, Crain, Jensen, Watson, Juhasz, Mandel, Liu, Mahon, Wilson, Lynch-Jonely, Leedom, and Fuller]{Feh25d}
Yao Fehlis, Charles Crain, Aidan Jensen, Michael Watson, James Juhasz, Paul Mandel, Betty Liu, Shawn Mahon, Daren Wilson, Nick Lynch-Jonely, Ben Leedom, and David Fuller.
\newblock Accelerating drug discovery through agentic {AI}: A multi-agent approach to laboratory automation in the {DMTA} cycle.
\newblock \emph{arXiv preprint arXiv:2507.09023}, 2025.
\newblock \doi{10.48550/arXiv.2507.09023}.

\bibitem[Gao et~al.(2025)Gao, Chang, Que, Xiong, Zhang, Qi, Liu, Wang, Ding, Li, Pan, Xie, Yan, Yan, and Zhang]{Gao25}
Jing Gao, Junhan Chang, Haohui Que, Yanfei Xiong, Shixiang Zhang, Xianwei Qi, Zhen Liu, Jun-Jie Wang, Qianjun Ding, Xinyu Li, Ziwei Pan, Qiming Xie, Zhuang Yan, Junchi Yan, and Linfeng Zhang.
\newblock {UniLabOS}: An {AI}-native operating system for autonomous laboratories.
\newblock \emph{arXiv preprint arXiv:2512.21766}, 2025.
\newblock \doi{10.48550/arXiv.2512.21766}.

\bibitem[Angelopoulos et~al.(2026)Angelopoulos, Cahoon, and Alterovitz]{Ang26}
A.~Angelopoulos, J.~F. Cahoon, and R.~Alterovitz.
\newblock From prompts to protocols: An {AI} agent for laboratory automation.
\newblock \emph{arXiv preprint arXiv:2605.16552}, 2026.
\newblock \doi{10.48550/arXiv.2605.16552}.

\bibitem[Munguia-Galeano et~al.(2026)Munguia-Galeano, Zhou, Veeramani, Fakhruldeen, Longley, Clowes, and Cooper]{munguia2026chemisteye}
Francisco Munguia-Galeano, Zhengxue Zhou, Satheeshkumar Veeramani, Hatem Fakhruldeen, Louis Longley, Rob Clowes, and Andrew~I. Cooper.
\newblock Chemist eye: a visual language model-powered system for safety monitoring and robot decision-making in self-driving laboratories.
\newblock \emph{Digital Discovery}, 5\penalty0 (5):\penalty0 2209--2220, 2026.
\newblock \doi{10.1039/d6dd00062b}.
\newblock URL \url{https://pubs.rsc.org/dd/article/5/5/2209/1238390/Chemist-Eye-a-visual-language-model-powered-system}.

\bibitem[Haddadnia et~al.(2026)Haddadnia, Sim, Hao, Hotta, Seifrid, Fedotov, Wu, Leong, Strieth-Kalthoff, and Aspuru-Guzik]{Had26}
Mohammad Haddadnia, Malcolm Sim, Han Hao, Kazuhiro Hotta, Martin Seifrid, Dennis Fedotov, Tony~C. Wu, Shi~Xuan Leong, Felix Strieth-Kalthoff, and Al{\'a}n Aspuru-Guzik.
\newblock {MoSeS} enables autonomous chromatography operation through programmable analytical expertise.
\newblock \emph{ChemRxiv}, 2026.
\newblock \doi{10.26434/chemrxiv.15008042/v1}.
\newblock August 30.

\bibitem[Ruiz-Soriano et~al.(2024)Ruiz-Soriano, Lamelza, Pizzamiglio, and Just-Baringo]{Rui24}
A.~Ruiz-Soriano, L.~Lamelza, E.~Pizzamiglio, and X.~Just-Baringo.
\newblock Synthesis of tetra-ortho-methoxylated azobenzene photoswitches via sequential catalytic {C--H} activation and methoxylation.
\newblock \emph{The Journal of Organic Chemistry}, 89:\penalty0 17141--17146, 2024.
\newblock \doi{10.1021/acs.joc.4c01554}.

\end{thebibliography}
\bibliographystyle{unsrtnat}
}

\clearpage

\appendix

\section{Experimental Procedure}
\label{app:experiment}
This appendix records the materials, analytical methods, and experimental procedures of the oxidant screening.

\subsection{Materials and analytical methods}

The authentic 4,4$^\prime$-dimethylazobenzene reference used to confirm the product assignment and to prepare the calibration standards was synthesized in-house from \textit{p}-toluidine by manganese-dioxide oxidation (10 equivalents of \ce{MnO2} in toluene at a substrate concentration of 0.1~M) following Ruiz-Soriano \textit{et al.}~\citep{Rui24}, as described in Section~\ref{sec:campaign}.

All LC--MS experiments were performed on an Agilent 1290 Infinity II UPLC (G7120A binary pump, G7167B multisampler, G7116B column thermostat, G7117B diode-array detector) coupled to an Agilent InfinityLab Pro iQ single-quadrupole mass spectrometer (G6160B) with electrospray ionization. Separations used an Agilent ZORBAX RRHD Eclipse Plus C18 column (2.1~$\times$~50~mm, 1.8~$\mu$m particle size; total length 50~mm), not thermostatted, with mobile phases of 0.10\% formic acid in water (A) and 0.11\% formic acid in acetonitrile (B) at a flow rate of 0.6~mL\,min$^{-1}$. The 7-minute gradient is given in Table~\ref{tab:app-uplc-method}; the injection volume was 0.2~$\mu$L unless otherwise stated. Diode-array signals were recorded at 210, 230, 254, 270, 290, 310, and 330~nm (1~nm bandwidth) alongside full UV--visible spectra. The mass spectrometer acquired positive-mode full scans over $m/z$ 85--1275 with a dynamic fragmentor ramp (100--170~V), a capillary voltage of 3500~V, drying gas at 325~$^\circ$C and 11~L\,min$^{-1}$, and a nebulizer pressure of 50~psi.

\begin{table}[h]
\centering
\small
\caption{UPLC gradient method (7.0~min total). Mobile phase A: 0.1\% formic acid in water; B: 0.1\% formic acid in acetonitrile.}
\label{tab:app-uplc-method}
\begin{tabular}{cccc}
\toprule
Time (min) & Flow (mL\,min$^{-1}$) & A (\%) & B (\%) \\
\midrule
0.0 & 0.6 & 95 & 5 \\
0.7 & 0.6 & 95 & 5 \\
5.0 & 0.6 & 5 & 95 \\
6.5 & 0.6 & 5 & 95 \\
6.6 & 0.6 & 95 & 5 \\
7.0 & 0.6 & 95 & 5 \\
\bottomrule
\end{tabular}
\end{table}

\subsection{Agent-generated experimental design}

The agent proposed an oxidant screen for the oxidative dimerization of \textit{p}-toluidine to 4,4$^\prime$-dimethylazobenzene. Table \ref{tab:app-stock-concentrations} shows the concentration of the prepared stock solutions. The design was generated from the target transformation and the reagents available in the self-driving laboratory. It compared hydrogen peroxide, sodium hypochlorite, Oxone\textsuperscript{\textregistered}, ammonium persulfate, potassium permanganate, and hypochlorite-mediated TEMPO and TEMPOL systems across four nominal oxidant loadings: 0.5, 1, 2, and 4 equivalents relative to \textit{p}-toluidine.

\begin{table*}[h]
\centering
\small
\caption{Concentrations of stock and working solutions used in the experiment. The \ce{NaOCl} concentration was not independently assayed.}
\label{tab:app-stock-concentrations}
\begin{tabular}{p{0.25\textwidth} p{0.20\textwidth} p{0.25\textwidth} p{0.20\textwidth}}
\toprule
Solution & Solvent & Concentration & Use \\
\midrule
\textit{p}-Toluidine stock & Acetonitrile & 20~mM & Reaction wells and G1--G3 \\
\ce{H2O2} stock & Water & 0.20~M \ce{H2O2} & Row A \\
\ce{NaOCl} stock & Water & 0.20~M \ce{NaOCl} & Rows B and F \\
Oxone\textsuperscript{\textregistered{}} stock & Water & 0.10~M Oxone\textsuperscript{\textregistered{}} (triple salt), 0.20~M active \ce{KHSO5} & Row C \\
Ammonium persulfate stock & Water & 0.20~M & Row D \\
\ce{KMnO4} stock & Water & 0.20~M & Row E \\
TEMPOL stock & Water & 20~mM & Row B and G2 \\
TEMPO stock & Water & 20~mM & Row F and G3 \\
Sodium thiosulfate solution & Water & 1.0~M & Quench \\
Acetanilide solution & Acetonitrile & 2.0~mM & Internal-standard solution \\
\bottomrule
\end{tabular}
\end{table*}

The design used 28 wells of an Agilent 2.0~mL deep-well plate (5043-9302) with 24 oxidation reactions and four controls. To each well 100~$\mu$L of a  20~mM \textit{p}-toluidine stock solution in acetonitrile was added. The oxidant and water volumes were then adjusted to give a total reaction volume of 200~$\mu$L. The resulting plate allowed the agent to compare oxidant identity and loading with a common liquid-handling pattern.

\subsection{Plate preparation}

Rows A--E contained the oxidant loadings shown in Table~\ref{tab:app-plate}. For these rows, the oxidant volumes were 5, 10, 20, and 40~$\mu$L, corresponding to 0.5, 1, 2, and 4 equivalencies relative to the limiting reagent \textit{p}-toluidine. Each well was diluted with water to bring the final reaction volume to 200~$\mu$L after addition of the 100~$\mu$L substrate stock solution. All additions were made by hand with single-channel pipettes: volumes of 20~$\mu$L or less with a P20 pipette and larger volumes with a P200 pipette (20--200~$\mu$L).

The control wells were prepared as shown in Table~\ref{tab:app-controls}

The plate was sealed after preparation with an Agilent PlateLoc Thermal Microplate Sealer and placed on a Torrey Pines shaker. The oxidant-screening reactions were run for 16~h at shaker setting 3 (on a shaker with levels ranging 0 to 9). The same procedure was used for the reaction wells and controls.

\begin{table*}[h]
\centering
\small
\caption{Composition of reaction mixtures. 100~$\mu$L of 20~mM p-toluidine stock solution was added to all reaction wells before the listed additions.}
\label{tab:app-plate}
\setlength{\tabcolsep}{4pt}
\begin{tabular}{>{\raggedright\arraybackslash}p{0.16\textwidth} *{4}{>{\raggedright\arraybackslash}p{0.185\textwidth}}}
\toprule
Row & 0.5 equivalents & 1 equivalent & 2 equivalents & 4 equivalents \\
\midrule
A: \ce{H2O2} & 5~$\mu$L \ce{H2O2} + 95~$\mu$L water & 10~$\mu$L \ce{H2O2} + 90~$\mu$L water & 20~$\mu$L \ce{H2O2} + 80~$\mu$L water & 40~$\mu$L \ce{H2O2} + 60~$\mu$L water \\
B: TEMPOL + \ce{NaOCl} & 5~$\mu$L \ce{NaOCl} + 10~$\mu$L TEMPOL + 85~$\mu$L water & 10~$\mu$L \ce{NaOCl} + 10~$\mu$L TEMPOL + 80~$\mu$L water & 20~$\mu$L \ce{NaOCl} + 10~$\mu$L TEMPOL + 70~$\mu$L water & 40~$\mu$L \ce{NaOCl} + 10~$\mu$L TEMPOL + 50~$\mu$L water \\
C: Oxone\textsuperscript{\textregistered} & 5~$\mu$L Oxone + 95~$\mu$L water & 10~$\mu$L Oxone + 90~$\mu$L water & 20~$\mu$L Oxone + 80~$\mu$L water & 40~$\mu$L Oxone + 60~$\mu$L water \\
D: ammonium persulfate & 5~$\mu$L persulfate + 95~$\mu$L water & 10~$\mu$L persulfate + 90~$\mu$L water & 20~$\mu$L persulfate + 80~$\mu$L water & 40~$\mu$L persulfate + 60~$\mu$L water \\
E: \ce{KMnO4} & 5~$\mu$L \ce{KMnO4} + 95~$\mu$L water & 10~$\mu$L \ce{KMnO4} + 90~$\mu$L water & 20~$\mu$L \ce{KMnO4} + 80~$\mu$L water & 40~$\mu$L \ce{KMnO4} + 60~$\mu$L water \\
F: TEMPO + \ce{NaOCl} & 5~$\mu$L \ce{NaOCl} + 10~$\mu$L TEMPO + 85~$\mu$L water & 10~$\mu$L \ce{NaOCl} + 10~$\mu$L TEMPO + 80~$\mu$L water & 20~$\mu$L \ce{NaOCl} + 10~$\mu$L TEMPO + 70~$\mu$L water & 40~$\mu$L \ce{NaOCl} + 10~$\mu$L TEMPO + 50~$\mu$L water \\
\bottomrule
\end{tabular}
\end{table*}

\begin{table}[h]
\centering
\small
\caption{Executed control reactions.}
\label{tab:app-controls}
\begin{tabular}{p{0.12\columnwidth} p{0.57\columnwidth} p{0.18\columnwidth}}
\toprule
Well & Composition & Purpose \\
\midrule
G1 & 100~$\mu$L \textit{p}-toluidine stock + 100~$\mu$L water & Substrate control \\
G2 & 100~$\mu$L \textit{p}-toluidine stock + 10~$\mu$L TEMPOL stock + 90~$\mu$L water & TEMPOL control \\
G3 & 100~$\mu$L \textit{p}-toluidine stock + 10~$\mu$L TEMPO stock + 90~$\mu$L water & TEMPO control \\
G4 & 100~$\mu$L acetonitrile + 100~$\mu$L water & Solvent blank \\
\bottomrule
\end{tabular}
\end{table}

\subsection{Quenching and internal-standard addition}

A sodium thiosulfate solution was prepared by dissolving 1.24~g sodium thiosulfate pentahydrate in 3.5~mL water and diluting to 5~mL in a volumetric flask. An acetanilide solution was prepared by combining 1.0~mL of 10~mM acetanilide stock with 4.0~mL acetonitrile in a 5~mL volumetric flask.

After the 16-hour reaction, the seal was removed from the whole plate at once, and 100~$\mu$L of the 1.0~M thiosulfate solution was added to every well by hand with an 8-channel P300 pipette from a reagent trough, dispensing one column at a time. The plate was then shaken on the Torrey Pines shaker at level 5 for about 1~min. Next, 100~$\mu$L of the acetanilide solution was added to every well in the same way with fresh tips, and the plate was shaken at level 3 for 5 minutes then sealed by hand with a sealing mat.  

\subsection{Manual drying and sample preparation}

After addition of internal standard, each well was filtered by hand through a pipette tip packed with magnesium sulfate into a new 96-well polypropylene plate (InfinityLab 96-well plate, 2.0~mL round U-shaped wells, 45~mm; Agilent part number 5043-9302) and washed with approximately 1.5~mL of acetonitrile.

The prepared plate was sealed with a silicone sealing mat (InfinityLab sealing mat, 96 round wells; Agilent part number 5043-9317) then placed into the LC autosampler for analysis.

\subsection{Verification of the product assignment and ranking}
\label{app:verification}

\begin{figure}[H]
\centering
\includegraphics[width=\textwidth]{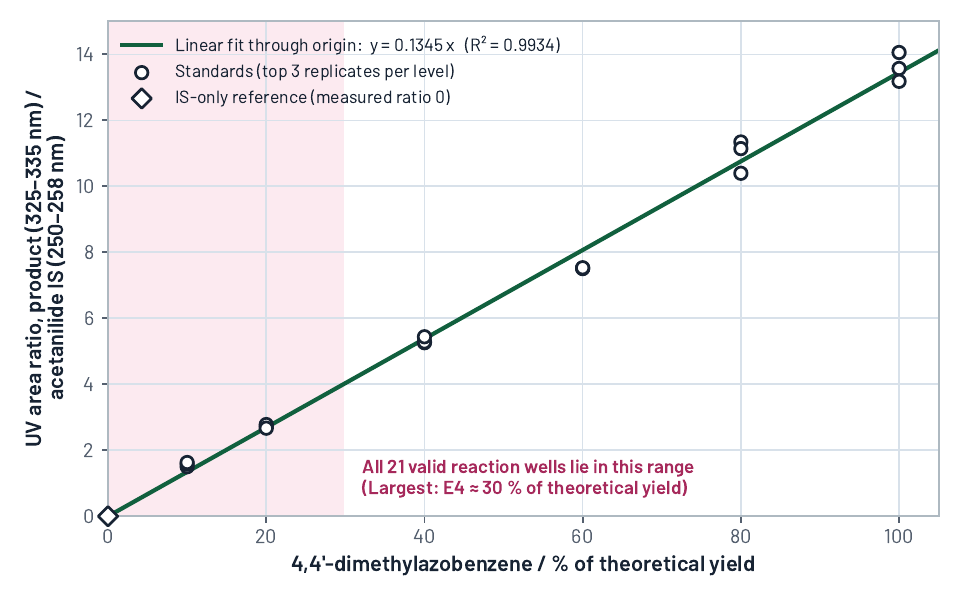}
\caption{UV internal-standard calibration for 4,4$^\prime$-dimethylazobenzene. A six-level calibration series of the synthesized standard (0.1--1.0~$\mu$mol, constant acetanilide internal-standard spike) was integrated with the same recipe as the reaction wells; the three most concordant replicates per level are shown (the 1.0~$\mu$mol level was re-prepared after its first-batch vials read low against both the on-plate standards and an independent earlier series; the re-run vials confirmed the expected response). The product amount is expressed as a percentage of a reaction well's theoretical full-conversion yield (1.0~$\mu$mol = 100\%), and the response as the ratio of the product band (325--335~nm) to the internal standard (250--258~nm). An acetanilide-only reference injection showed no detectable product and is plotted at the origin (open diamond); pure-acetonitrile blanks likewise showed no signal at the product's retention time. The fit is forced through the origin and its statistics use the standards only. The shaded band marks the range spanned by the 21 valid reaction wells, with amounts obtained from the on-plate replicate of the standards: the best condition, E4, reached approximately 30\% of the theoretical yield.}
\label{fig:app-calibration}
\end{figure}

Three post-campaign checks were performed. First, the UV (total-wavelength DAD) product and internal-standard peaks of the campaign chromatograms were integrated in the vendor chromatography data system (Agilent OpenLab CDS, version 2.8 Update 09 with Feature Pack 02); this placed \ce{KMnO4} far above the other oxidants with E4 as the best well, while the best wells of Oxone\textsuperscript{\textregistered}, TEMPO/\ce{NaOCl} and TEMPOL/\ce{NaOCl} differed by less than 10\% (7.4--8.2\% of E4), too close for this readout to order (Table~\ref{tab:app-integration-comparison}). Second, an authentic sample of 4,4$^\prime$-dimethylazobenzene was synthesized by manganese-dioxide oxidation of \textit{p}-toluidine; it reproduced the chromatographic and mass-spectral signature of the product feature, confirming the assignment. Third, the synthesized standard was used to calibrate the product's UV response as the ratio of the product band (325--335~nm) to the acetanilide internal standard (250--258~nm): a dedicated six-level calibration series (0.1--1.0~$\mu$mol) established the response (Figure~\ref{fig:app-calibration}), and an on-plate replicate of the standards (five levels, 0.2--1.0~$\mu$mol, spiked with the same acetanilide amount as the reaction wells, acquired on the same plate and LC method) anchored the transfer to the wells. Every campaign well was then re-integrated with the same recipe, both as the raw product area and as the product-to-internal-standard ratio; the ratio cancels the variable dilution introduced by the manual filtration transfer, which raw peak areas do not. This absorbance-based readout identified E4 as the best condition; the best wells of the two mid-ranked systems (Oxone\textsuperscript{\textregistered} and TEMPO/\ce{NaOCl}) were within 5\% of each other, inside this readout's replicate scatter. Converted through the on-plate curve, the two strongest wells fall within the calibrated range, with the best condition (E4) at approximately 30\% of the theoretical yield; every other well lies below the lowest standard, so the screen as a whole is reported as a relative ranking.

\begin{table}[H]
\centering
\small
\caption{UV product response of each reaction well from the agent's analysis and from the vendor-software integration, each expressed as the product-to-internal-standard ratio relative to well E4 (E4 = 100). \emph{Agent, UV}: the absorbance-based readout of the third check (product band 325--335~nm over the internal standard 250--258~nm), as plotted in Figure~\ref{fig:screen}; its relative uncertainty is about 13\%. \emph{OpenLab CDS, UV}: product and internal-standard peaks of the total-wavelength DAD signal integrated in Agilent OpenLab CDS. n.d., not detected (for OpenLab CDS, below the vendor integration threshold); ---, no internal-standard peak (failed transfer) or low-quality draw (F2, agent UV).}
\label{tab:app-integration-comparison}
\begin{tabular}{llccc}
\toprule
 & & & \multicolumn{2}{c}{Product response (\% of E4)} \\
\cmidrule(lr){4-5}
Well & Oxidant & Equiv. & Agent, UV & OpenLab CDS, UV \\
\midrule
A1 & \ce{H2O2} & 0.5 & 0.57 & n.d. \\
A2 &  & 1 & n.d. & n.d. \\
A3 &  & 2 & n.d. & n.d. \\
A4 &  & 4 & n.d. & n.d. \\
\addlinespace
B1 & TEMPOL/\ce{NaOCl} & 0.5 & 2.4 & n.d. \\
B2 &  & 1 & --- & --- \\
B3 &  & 2 & 2.9 & 3.5 \\
B4 &  & 4 & 6.8 & 7.4 \\
\addlinespace
C1 & Oxone\textsuperscript{\textregistered} & 0.5 & 5.3 & n.d. \\
C2 &  & 1 & 8.2 & 8.0 \\
C3 &  & 2 & 8.2 & 8.2 \\
C4 &  & 4 & --- & --- \\
\addlinespace
D1 & Ammonium persulfate & 0.5 & 0.65 & n.d. \\
D2 &  & 1 & 0.77 & n.d. \\
D3 &  & 2 & 1.0 & n.d. \\
D4 &  & 4 & 1.1 & n.d. \\
\addlinespace
E1 & \ce{KMnO4} & 0.5 & 30.2 & 33.6 \\
E2 &  & 1 & 77.6 & 85.2 \\
E3 &  & 2 & 27.0 & 27.0 \\
E4 &  & 4 & 100 & 100 \\
\addlinespace
F1 & TEMPO/\ce{NaOCl} & 0.5 & 2.5 & n.d. \\
F2 &  & 1 & --- & n.d. \\
F3 &  & 2 & 6.3 & n.d. \\
F4 &  & 4 & 7.9 & 7.5 \\
\bottomrule
\end{tabular}
\end{table}

The vendor-software integration and the agent's absorbance-based integration measured the same peaks. For the nine wells in which both reported the product, the ratio of the OpenLab CDS area to the agent's area was constant to within 7.4\% for the product peak (coefficient of variation 5.3\%), and to within 11.2\% for the internal-standard peak across all 21 wells in which both reported it (coefficient of variation 4.7\%); the absolute scale differs because OpenLab CDS integrated the total-wavelength DAD signal, whereas the agent integrated compound-specific bands (325--335~nm for the product, 250--258~nm for the internal standard). The product-to-internal-standard ratios from OpenLab CDS were systematically lower than the agent's, by a median of 14\% (range $-18$\% to $+3$\%), consistent with this difference in spectral weighting between the two signals; because this offset is shared across wells, it does not change the ordering within any row. The two methods differ mainly in sensitivity: the OpenLab CDS peak table reported no product peak below about 8\% of the E4 product area, whereas the agent's integration recovered smaller peaks, including C1 and F3 (6--8\% of E4) and wells in rows A, B, D and F at about 1\% of E4. The ``n.d.'' entries in the OpenLab CDS column of Table~\ref{tab:app-integration-comparison} therefore mark peaks below the vendor integration threshold rather than the absence of product.

\section{Closed-Loop Campaign Workflow}
\label{app:loop}

This appendix describes the workflow used after the quenched plate was placed in the autosampler. It covers the system components, campaign authorization, the four-stage adaptive loop, and offline evaluation against archived measurements.

\subsection{System components and authority}

The deployed workflow lives in the SDL net (Figure~\ref{fig:network}). The instrument PC runs the chromatography data system and a sidecar service that exposes instrument status, cooperative claims, a first-in-first-out run queue, an allow-list of approved acquisition methods, and abort control. The sidecar uploads completed raw files and updates the corresponding measurement record.

The data server provides the project-scoped experiment record store, including experiments, samples, measurements, files, analyses, and plans. The laboratory server hosts the agent service, project rooms, tools built with Python LC--MS analysis libraries, campaign tools for the agent to plan measurements, and the watcher that connects completed measurements to new agent turns. Runtime state and tool-call traces are stored with the campaign ledger, while result files are staged in the room workspace. User PCs access the project room through an authenticating edge.

The agent communicates with the instrument PC directly via state-changing requests that pass through the SDL net, where project scope, instrument ownership, queue capacity, approved methods, and device state are checked. Thus, the agent can propose a measurement but cannot bypass the execution controls.

\subsection{Registration and campaign approval}

The plate is registered as an experiment with one sample per well, including its position, reagent system, loading, and notes. Only registered wells can be selected for measurement. The human-approved campaign execution detail then specifies the objective, candidate wells, metric, acquisition method, injection volume, and maximum number of injections. The method is checked against the instrument run contract before the campaign can start.

The execution detail is initially presented to the human as an approval card (Appendix \ref{app:workspace}.3). After approval, every submission is validated against the fixed set of criteria rather than regenerated by the agent. A room-level \emph{Agent execution} control gates autonomous operation, and an emergency stop refuses new submissions and pauses the watcher. These controls remain outside the agent's decision process.

\begin{figure}[t]
\centering
\includegraphics[width=\textwidth]{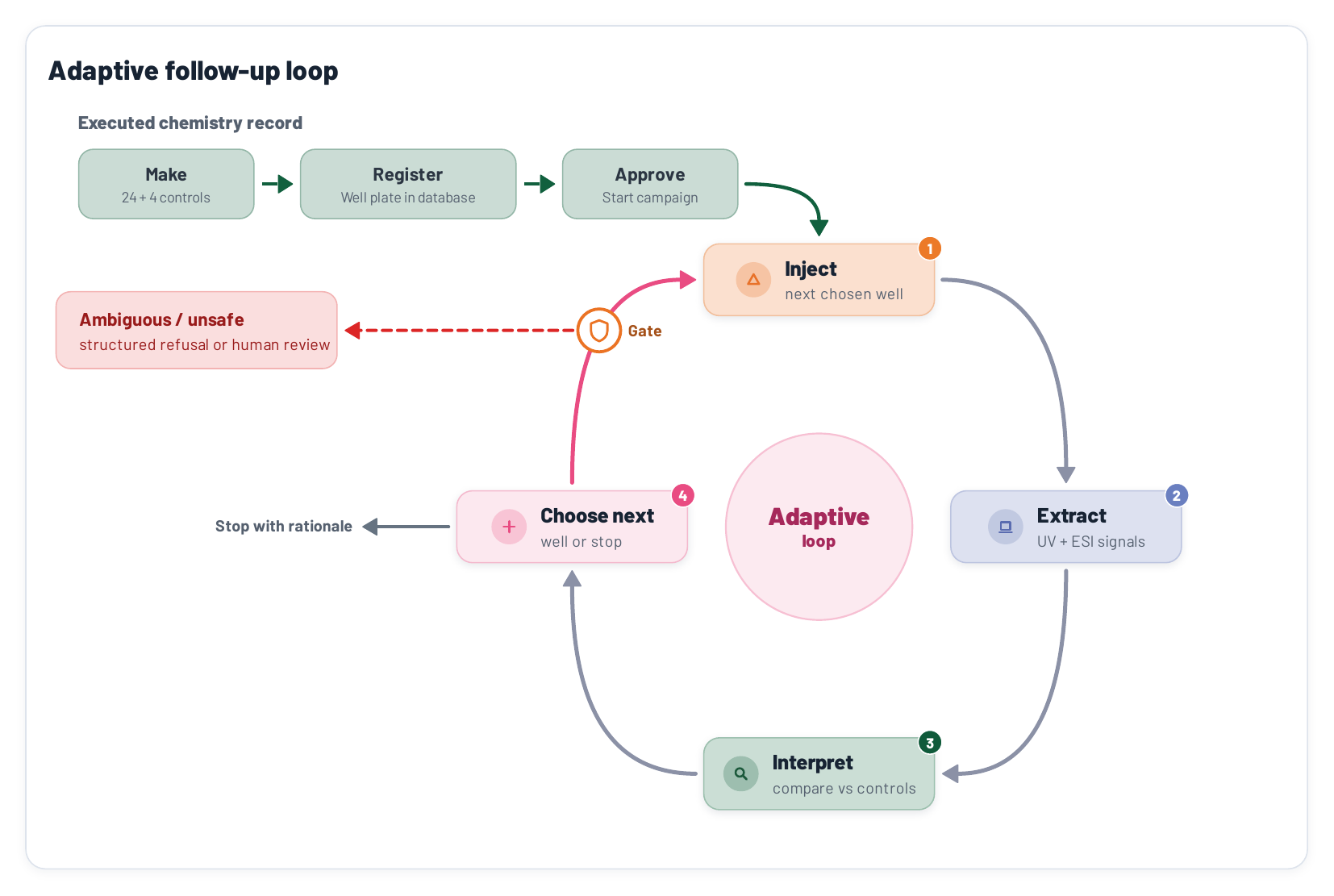}
\caption{The agentic loop during execution of the oxidant screening campaign.}
\label{fig:loop-execution}
\end{figure}

The adaptive loop has four stages, distinct from the one-time preparation steps of registration, incubation, and filtering shown in Figure~\ref{fig:loop-execution}

\subsection{One closed-loop iteration}

\begin{enumerate}
\item \textbf{Inject.} The agent reads the objective, campaign state, per-well results, quality flags, and remaining budget, then selects one registered well. The submission path checks that the campaign is active, the well is allowed and not already in flight, the budget remains, and the analytical method matches the approved method. The sidecar then obtains the instrument claim, runs the approved method, and uploads the completed raw file to a new measurement record.

\item \textbf{Extract.} The watcher detects the completed measurement, stages the file in the project room, updates the ledger, and starts a new agent turn. Deterministic tools load the file and extract the diagnostic ion chromatograms, peaks, and other signals needed for analysis. Because the turn is reconstructed from persistent records, the loop can resume after a service restart.

\item \textbf{Interpret.} The agent interprets the extracted signals, computes the campaign metric, and records the result, notes, and quality flags. Failed injections or measurements with inadequate internal-standard response can be remeasured or excluded with an explicit note.

\item \textbf{Choose next.} The agent selects another well, requests a justified remeasurement, ends the campaign when the objective is answered, or stops when the approved budget is exhausted. The final summary, ranking, and closed ledger remain in the project record.
\end{enumerate}

\subsection{Retrospective evaluation}
\label{app:loop:replay}

The retrospective evaluation runs the same agent, tools, watcher, submission criteria, and validation path as the physical campaign. Only the instrument interface is replaced by a simulator. Each submitted well is mapped to its autosampler position, and the simulator returns the corresponding LC--MS file from the exhaustively acquired plate. The file then completes the same measurement-record and validation steps as a physical measurement.

This setup preserves the interaction between agent decisions, measurement results, quality checks, and budget constraints while using a fixed reference dataset. Each campaign records its tool trace and ordered list of submitted wells. Performance is evaluated from the resulting ledger: whether the true best condition was measured and reported, how many injections were used, and when the optimal condition was first identified (Figure~\ref{fig:budget}).

\section{Simulated Instrument}
\label{app:simlab}

The simulated instrument provides an instrument-level substitute for the LC--MS device, allowing the workflow in Appendix~\ref{app:loop} to run without controlling a physical instrument. The agent, analysis tools, campaign envelope, record contracts, and validation path remain the same as in the deployed system; the LC--MS acquisition and instrument-control layer is replaced. The evaluation harness supports two use cases: automated evaluation of closed-loop behavior with synthetic or archived measurements, and interactive evaluation with archived measurements through the deployed interface.

\subsection{Architecture}

The simulated-instrument evaluation harness contains four principal components.

\begin{itemize}
\item \textbf{Record store.} An in-memory implementation follows the production record-store schema, identifier conventions, and plan state transitions. A seeded clock and deterministic identifiers make repeated runs produce reproducible histories. The implementation sits beneath the record client, so the agent exercises the same schemas and state checks without a network service.

\item \textbf{Simulated instrument and queue.} The simulated instrument enforces the same claim discipline as the laboratory device: one claim holder, token-gated control operations, and a first-in-first-out queue. Queue capacity and claim conflicts are therefore explicit test conditions rather than timing accidents.

\item \textbf{Execution backend and completer.} The simulated submission path mirrors the production route. It verifies the sample, creates a submitted measurement record, acquires the simulated instrument claim, queues the job, and releases the claim. The completer removes queued jobs, obtains a result from a configured synthesizer, registers the result file, and completes the measurement record. From the watcher and agent's perspective, this follows the same completion path as a physical acquisition.

\item \textbf{Result sources.} The instrument simulator supports kinetic and screening simulations, as well as evaluations based on archived experimental measurements. Kinetic and screening simulations generate synthetic files from a configured response model; archived-measurement evaluations map each well to the real LC--MS file obtained during exhaustive acquisition. Synthetic files use the production file format and pass through the production loading and analysis pipeline.
\end{itemize}

The evaluation harness can run entirely in process for headless experiments. In the deployed service, setting \texttt{CAMPAIGN\_EXECUTION=sim} replaces only the instrument: campaign records, files, plans, and registration still use the real backend, while the simulated queue and completer provide acquisition. This preserves the boundary between the instrument plane and the record plane.

\subsection{Automated evaluation}

The evaluation harness runs replicated campaigns with the production agent against the simulated instrument. Each replicate follows the live sequence:

\[
\text{agent turn} \rightarrow \text{instrument queue} \rightarrow \text{simulated acquisition} \rightarrow \text{watcher event} \rightarrow \text{next agent turn}.
\]

For the archived-plate evaluation, a manifest maps each registered well to its latest valid LC--MS acquisition at the target volume, or to its latest failed acquisition when no valid file exists. The agent receives files only for wells it selects; no new chemical measurements are generated. Each campaign records the tool trace, injection budget, ordered submitted wells, and per-well ledger. Performance is evaluated by measurement economy relative to exhaustive acquisition, file-to-well attribution, agreement with the fixed reference outcome, and whether the best condition was measured and reported (Figure~\ref{fig:budget}). This evaluates decision quality and workflow behavior against fixed experimental measurements rather than fresh chemical discovery.

\subsection{Interactive evaluation with archived measurements}

The archived-measurement evaluation can also be accessed through the deployed interface. A registered plate is paired with a measurement manifest and a campaign budget below the plate size. The dashboard, chat, Plate tab, and Loop tab then show the agent selecting a well, submitting a simulated instrument run, receiving the corresponding previously acquired LC--MS file, and selecting the next well (detailed information on the user interface is described in Appendix \ref{app:workspace}). Because interactive evaluation writes to the real record store, it is run under a throwaway project and does not contact the physical instrument. It evaluates the deployed workflow and decision process; it does not constitute a new experiment or add new chemical measurements.

\section{User Interface and Agent Capabilities}
\label{app:ui-agent}

This appendix summarizes the interface used to supervise a campaign and the capabilities available to the agent within it.

\subsection{Project room}

The interface is organized around a project room. An account banner identifies the signed-in laboratory user and limits access to that user's project memberships. The chat panel contains the conversation with the agent, while the workspace provides tabs for project guidance, individual instrument submissions, user data, plate registration, files, tool usage, workflow state, and campaign status.

The room header exposes the controls that govern autonomous operation: the model selector, the \emph{Agent execution} toggle, and the emergency stop. During a campaign, the header also shows the campaign state, injections used and remaining, wells completed or in flight, failures, and the current instrument phase. The workspace remains available when chat is collapsed, allowing a user to inspect records and campaign state without initiating an agent turn.

\subsection{Agent capabilities}

The agent is a language-model orchestrator; deterministic tools perform data access, analysis, and state changes. Its available tools are registered per room and are grouped as shown in Table \ref{tab:toolsets}.

\begin{table}[h]
\footnotesize
\setlength{\tabcolsep}{4pt}
\caption{Principal tool groups available to the agent.}
\label{tab:toolsets}
\begin{tabular}{@{}p{0.22\columnwidth}p{0.43\columnwidth}p{0.25\columnwidth}@{}}
\toprule
Tool group & Capability & Representative operations \\
\midrule
Analysis & LC--MS data loading, feature extraction, peak analysis, and quantification & load files, extract chromatograms, find peaks, inspect spectra, build tables \\
Records and workspace & Project-scoped records, files, computation, and trace export & read/write records, manage files, run sandboxed Python, export traces \\
Campaign & Defined workflows, such as plate registration and the closed-loop measurement workflow & draft/register plate, start campaign, read state, submit well, record result, end or abort \\
Instrument & Status inspection and direct single-injection workflow & check status, fill submission form, submit run \\
Grounding & Literature retrieval and scientific context & search literature, inspect metadata, retrieve open-access links \\
\bottomrule
\end{tabular}
\end{table}

There is no hand-coded acquisition policy in the campaign tools. During an approved screening campaign, the language model selects the next well from the objective, registered conditions, campaign ledger, quality flags, and remaining budget (Appendix~\ref{app:loop}).

\subsection{Authority and human controls}

The agent can inspect project-scoped records and instrument status, prepare plate and injection forms, search the literature, analyze staged data, write results and notes, and operate a campaign after its operation has been approved. All record operations are attributed to the room owner and the agent identity.

The \emph{Agent execution} toggle grants two additional permissions: autonomous plate registration and submission of a direct single-injection run. Within an approved campaign, the agent may submit registered wells, receive and analyze results, record metrics, remeasure failed wells, and close the campaign.

The agent cannot approve the campaign, access projects outside the owner's scope, run an unregistered sample, select a well outside the approved, exceed the injection budget, bypass an instrument claim, delete records, or alter its own tool-call history. A direct injection submitted while the toggle is off is refused; the agent can still prepare the form for human submission.

Four human controls govern the system:

\begin{itemize}
\item The \emph{Agent execution} toggle can be revoked at any time and takes effect before the agent's next state-changing action.
\item The campaign requires one-time human approval of the objective, method, candidate wells, injection volume, and budget.
\item The \emph{Stop} control cancels the current agent turn. A campaign can then be abandoned, while an injection already running on the instrument is allowed to finish and is recorded.
\item The \emph{E-STOP} revokes autonomous execution, refuses subsequent submissions, and freezes campaign wake-ups until deliberately released. It does not interrupt an injection already controlled by the instrument.
\end{itemize}

Tool calls, arguments, results, campaign state, and final summaries are retained in the project record. They are visible in the Tool Usage and Workflow panels, exportable as an ordered trace, and available for replay or comparison with a later execution.

\section{Prompts}
\label{app:prompts}

This appendix explains the prompts used to guide the agent. 

Two kinds of text govern a campaign. The first is fixed per deployment and not specific to this study: a persona system prompt and per-tool-set instructions. The second is written per campaign and is what actually steers the loop: the campaign details the human approves (for the live run) or the objective given to the harness (for the offline evaluation). This appendix describes the first briefly and reproduces the second verbatim.

\subsection{System prompt and tool instructions}

For the oxidant screening campaign example used, the agent's system prompt defines a single analytical-chemistry persona for LC--MS, GC--MS, and NMR data and states the division of labour in one sentence: for ad-hoc analysis the human has already run the instrument; inside an approved campaign the agent also executes, through the audited campaign tools, and there is no other path for it to start a run. It then lists what the agent can do per technique, how to respond to an analysis request (identify the data type, locate the file, call the workflow, reason about the result, say when it is inconclusive), which file to analyze when several are present, how to correlate UV and MS peaks given the inter-detector delay, and a few plotting conventions. Each registered tool set adds its own instructions: the campaign tool set carries the loop protocol (register~$\to$~approve~$\to$~submit one well per turn~$\to$~be woken~$\to$~analyze and record~$\to$~stop), the method-locking rule and its rationale (areas and retention times are only comparable within one method), the injection-volume cap, method-design and scouting practice, and the instruction that a refusal is information to correct from rather than an error; the LC--MS tool set describes the atomic extract~$\to$~detect~$\to$~fit~$\to$~identify chain and the handles that connect its steps; the grounding tool set describes literature search and trace export. None of this text names the oxidant screen or any strategy for it.

\subsection{Prompt used for oxidant screening}

Assembled per campaign as: objective $\cdot$ arm text $\cdot$ budget line $\cdot$ method text $\cdot$ autonomy clause $\cdot$ answer format. The reported sweep used the \emph{adaptive} arm, whose arm text is empty; the \emph{guided} arm's text follows for comparison.

{\small\begin{verbatim}
Start a screening campaign on plate {plate} (experiment {experiment_id}, already
  registered). Rows A-F are oxidant systems for the oxidative dimerization of
  p-toluidine - read each well's system and loading from the campaign state - and
  columns 1-4 are ascending nominal loadings (5/10/20/40 uL of that row's oxidant
  stock). Row G holds controls: G1 substrate-only (no oxidant), G2 TEMPOL-only, G3
  TEMPO-only, G4 solvent blank. Objective: find PRODUCT FORMATION, not reactant
  depletion. For each oxidant system, identify the loading that produces the most
  4,4'-dimethylazobenzene, and rank the systems by their best measured product
  formation. If the m/z 136 internal-standard area is below 340000, treat the 
  injection as invalid: retry the well once, and if it fails again report it as 
  unmeasurable and exclude it.
  Do not convert invalid or missing values into product_ratios. Minimize
  injections and do not measure wells whose outcome is already established by
  valid data.

Budget: <b> injections.

Your final message must end with one fenced ```json block of exactly this shape:
  {"background_ratio": <your measured G1 product_ratio>, "rows": {"<row letter>":
  {"system": "<name>", "best_product_ratio": <highest valid product_ratio in the
  row, or null>, "best_col": <column number of that maximum, or null>,
  "lowest_product_positive_col": <lowest column clearly above background, or
  null>}, ... one entry per system row}, "ranking": ["<row letter>", ...] (system
  rows ordered by best product_ratio, most product first)}.
\end{verbatim}}

UV detection was not used for the metric in the prompt or for the reference values. The samples contained approximately 86\% acetonitrile, whereas the LC gradient started at 5\% acetonitrile (Table~\ref{tab:app-uplc-method}); this solvent mismatch distorted the peaks of early-eluting compounds such as the substrate, so their UV areas could not be quantified reliably. The product eluted later and showed a characteristic \emph{trans}-azobenzene band at $\lambda_{\max}$ 332\,nm, but at the time of the campaign its UV response had not been validated for comparison across wells: a non-azo feature near 230\,nm co-eluted with the product, and without an authentic product standard the 332\,nm response could not be shown to be proportional across oxidant rows. The MS product-to-internal-standard ratio was therefore used in both the prompt and the fixed reference values. After the campaign, a UV calibration against the synthesized standard served as an independent check of the ranking (Appendix~\ref{app:verification}).

\subsection{The wake-up message}

Posted by the watcher into the room, one message per tick covering every newly completed or failed well; each one starts a fresh agent turn. Example from the live campaign:

{\small\begin{verbatim}
[campaign <campaign id>]
well <well location> (<well content>) complete
  -> catalog/<file name>.npz
Analyze new results, record them, and continue the campaign; decide whether to
  re-measure any failed well.
\end{verbatim}}

\section{The workspace interface}
\label{app:workspace}

This appendix walks through the interface in the order a campaign uses it. Each subsection names what the human does there and what the surface shows while the loop runs.

\subsection{Rooms and chat}

\begin{figure}[htbp]
\centering
\includegraphics[width=0.8\textwidth]{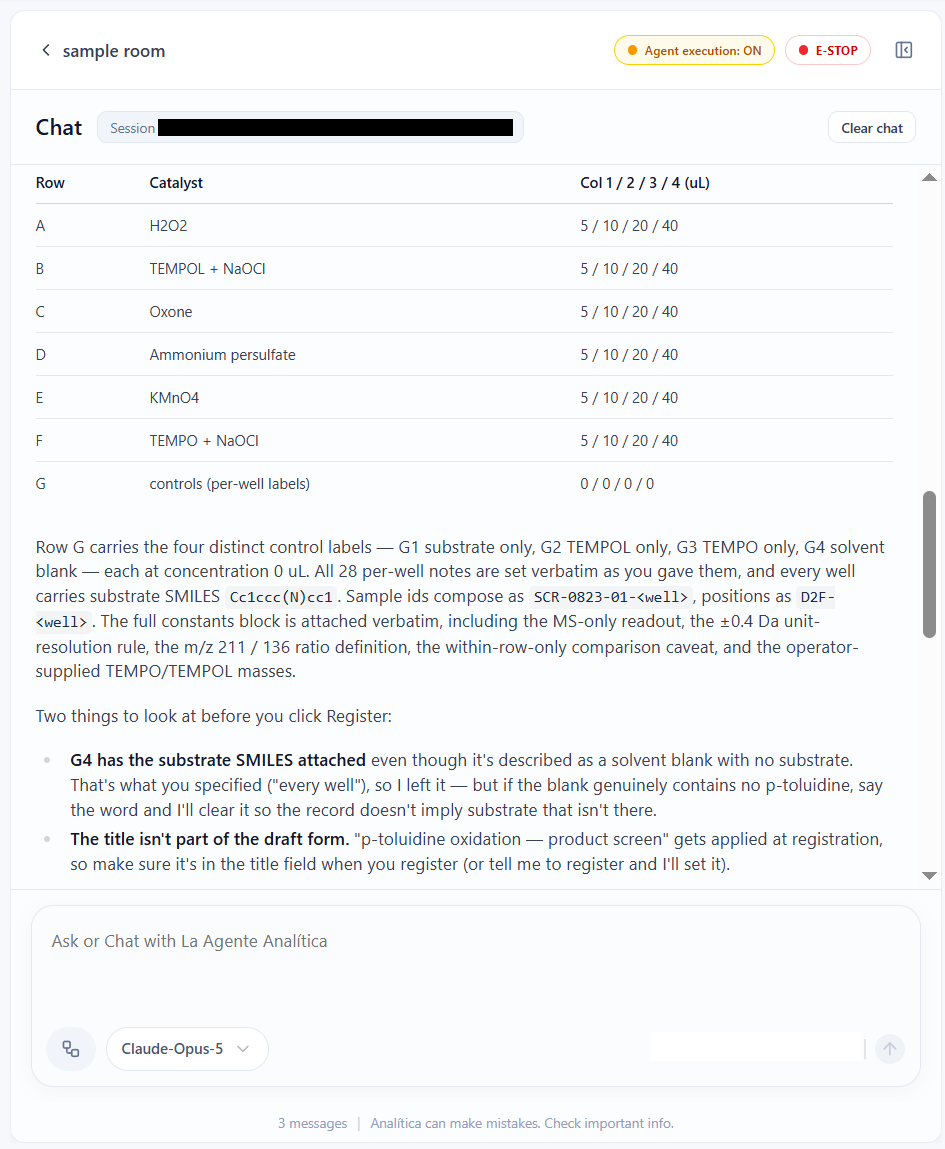}
\caption{The chat room while the oxidant screen is set up: the header with the Agent-execution toggle and E-STOP, the agent's proposed plate layout and notes for the human, and the model selector.}
\label{fig:room}
\end{figure}

A room is one conversation bound to a workspace directory, a model, and its own controls; rooms are listed in the left panel and can be switched or created at will, and the chat/workspace divider is draggable. The chat header holds the controls described in Appendix~\ref{app:ui-agent}: the model selector, the \emph{Agent execution} toggle, the \emph{E-STOP}, and, once a campaign is active, the status strip with information on the state, injections used against budget, wells done / in flight / failed (per-well metrics on hover), and a live narration of the instrument phase joined from the instrument's own queue (``A4 running on instrument $\cdot$ 12~m''), so the quiet minutes between a submission and its wake-up still show progress. The transcript shows every tool call the agent makes; \emph{Stop} cancels the current turn. Clearing the chat resets the agent session and its reasoning log but preserves the workspace files; the campaign's actions survive regardless in the ledger and the record store. Figure~\ref{fig:room} shows the agent chat room.

\subsection{Plate tab}

\begin{figure}[htbp]
\centering
\includegraphics[width=\textwidth]{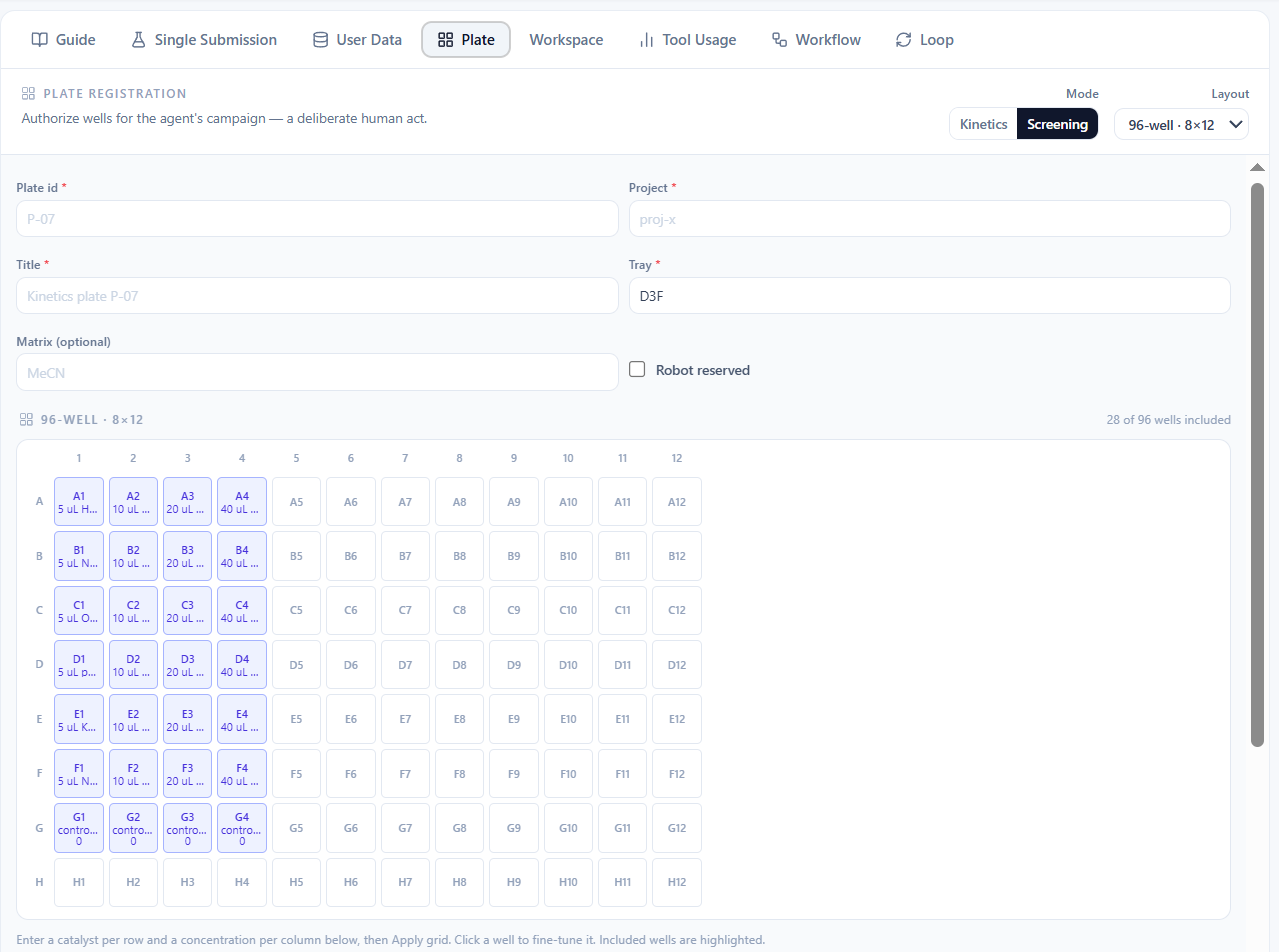}
\caption{The Plate tab while the oxidant screen is being registered: screening mode, the tray, and the 28 wells of the 96-well layout that carry the agent-proposed conditions.}
\label{fig:plate}
\end{figure}

Registration happens here (sign-in required). The human picks a layout from the library (96-well by default; a 32-vial 4$\times$8 layout is also available) and the tray it sits in, from which every well's autosampler position derives; sets how many wells carry sample; ticks \emph{robot-reserved} when the tray is the robot's; chooses the mode---\emph{kinetics} (a quench time per well) or \emph{screening} (a reagent per row, a loading per column, plate-level constants in plain language, and an optional note per well); and clicks \emph{Register plate}, which creates the experiment and one sample per well in the experimental record database. The agent can pre-fill either shape from a chat request and edit only the wells named in follow-up requests; the human corrects and confirms. Figure~\ref{fig:plate} shows the Plate tab while the oxidant screen is being registered. Only registered wells can ever be run. During a campaign the same grid is the live dashboard: wells colour by status (queued / done / failed) and show the recorded metric as a heat map as numbers land, read from the ledger snapshot the loop writes (\texttt{runs/<campaign>/campaign.json}).

\subsection{Campaign approval card}

\begin{figure}[htbp]
\centering
\includegraphics[width=0.8\textwidth]{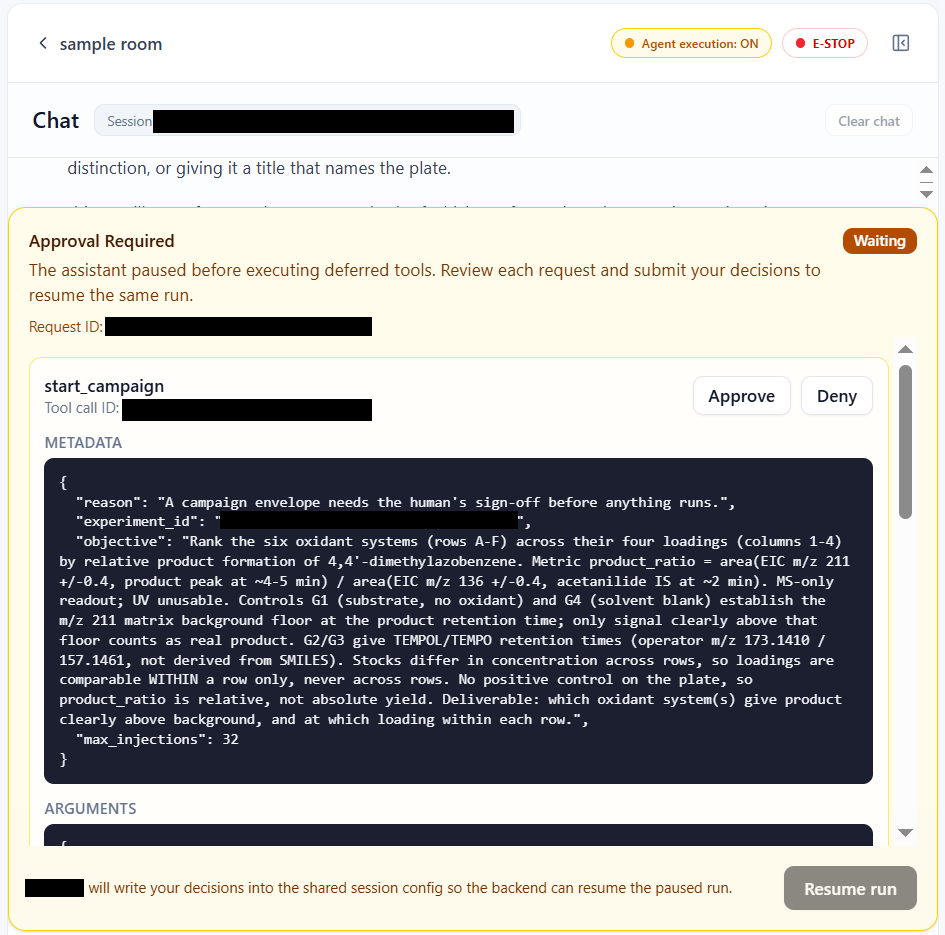}
\caption{The human must approve the campaign envelope before the agent can execute on the instrument.}
\label{fig:approval}
\end{figure}

When the agent calls \texttt{start\_campaign}, an Approve/Deny panel appears in the conversation showing the proposed execution plan: the candidate wells, the objective, the injection budget, and the method (a custom gradient stated in the prompt is baked into the method template and validated against the instrument contract \emph{before} the panel appears, so a broken method never reaches the queue). Approving is the single human sign-off; the proposal is recorded as a plan whose status tracks the campaign (\emph{draft}~$\to$~\emph{approved}~$\to$~\emph{executing}~$\to$~\emph{completed}, or \emph{abandoned}). Figure~\ref{fig:approval} shows the approval card for the oxidant screen: the wells it may touch, the objective text the agent derived from the operator's message, the budget, and the locked method. Until the human clicks \emph{Approve}, nothing can be submitted; after it, the loop runs without further sign-off and every submission is validated against this verified criteria instead of asked about. Denying leaves the plan as a draft and nothing reaches the queue; the agent can revise the campaign and ask again.

\subsection{Loop tab}

The Loop tab draws the closed loop as a ring (instrument run~$\to$~agent analyzes~$\to$~agent picks the next well~$\to$~back to the instrument) with the live station lit and labelled: which well, the instrument phase, elapsed time, and the tools running right now; hovering a station shows its detail. Everything drawn is derived from the campaign ledger, the instrument queue, and the tool-call log, not from parallel bookkeeping, so what the ring shows is what the records say. During the oxidant screen the ring was the quickest way to see where an injection was (queued, acquiring, staged, or being analyzed) without reading the transcript, and the tools listed at the analysis station (extracting ion chromatograms, detecting peaks, recording the metric) are the same calls that appear in the Tool Usage log. When the campaign ends the ring settles on the final state and the status strip above the chat reports the closing summary.

\subsection{Instrument tab}

\begin{figure}[htbp]
\centering
\includegraphics[width=\textwidth]{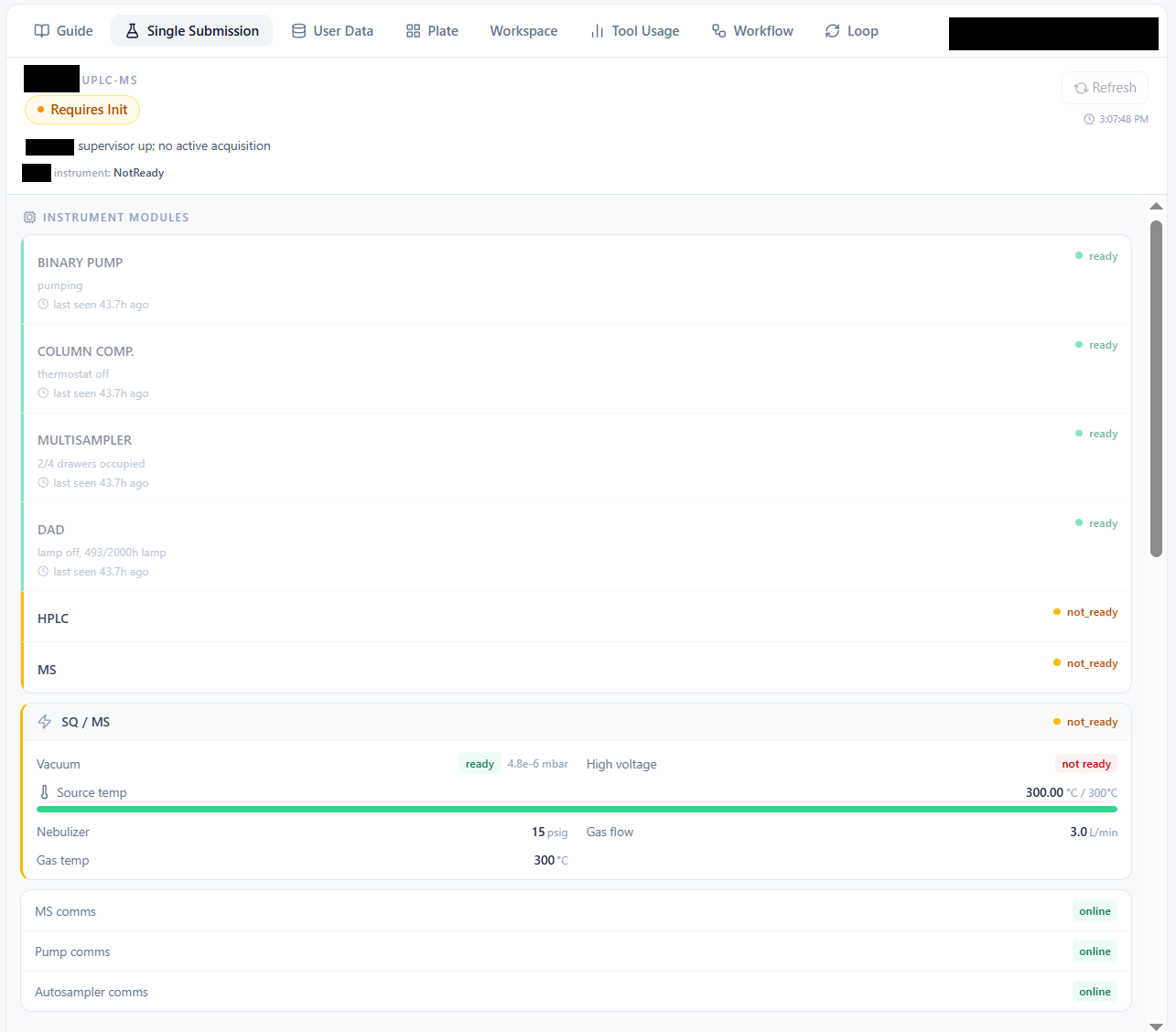}
\caption{The Single Submission tab shows live instrument status, module health, and the run queue, through which the human can also submit a run directly.}
\label{fig:instrument}
\end{figure}

The instrument panel (Figure~\ref{fig:instrument}) is the human's direct path to the LC--MS and the place to watch the queue. Its status header is driven by the raw instrument state (Ready, Busy---including runs started directly in the vendor software, Requires Init, Paused, Error), with required operator actions and the last fault; module rows show pump, column thermostat, multisampler, diode-array detector, and the LC and MS roll-ups, plus live MS sensor readings (vacuum, high voltage, source temperature, nebulizer and drying gas); consumable bars show solvent and waste levels once a run has populated the solvent log. The run queue lists the active job, pending jobs in order, and recent done or failed jobs with any instrument message. The submission form takes a registered sample identifier and project, the job definition, and \emph{Add job to queue}; validation runs locally, then the backend creates the linked records, claims the instrument, and queues the job. The agent can pre-fill this form (\texttt{draft\_hplc\_submission}) and, only with the toggle on, submit through the same route.

\subsection{User Data tab}

A three-level browser over the laboratory's record store scoped to the projects the identity may read: experiments grouped by project (agent-attributed rows are flagged), the samples of an experiment (a plate campaign shows one row per well with its condition and position), and the files of a sample. Selected files are sent to the room's \texttt{catalog/} directory (idempotently; a whole plate can be loaded and selected at once), which is also where the watcher stages campaign results, so a colleague in another room can pull the same plate's files and analyze them. File bytes come from the record store's read-only preview service, which enforces per-project access on every request.

\subsection{Workspace explorer, Tool Usage, and Workflow}

The Workspace tab is the room's file tree: uploads, agent outputs under \texttt{temp/}, staged results under \texttt{catalog/}, the ledger snapshot under \texttt{runs/}, exported traces under \texttt{traces/}, and a recycle bin; files can be previewed in place (data-file heat maps, molecules, images). The Tool Usage tab shows a bar chart of calls per tool and an expandable log of every call with its input and output; asking the agent to export the trace writes the same information, paired with results and timestamps from the durable store, to a JSON artifact. Together these three tabs are the human's view of what the agent did and why: the files it produced, the calls it made, and whether the same calls reproduce the same outputs.

\end{document}